\PassOptionsToPackage{numbers,compress}{natbib}
\documentclass{article}

\usepackage{iclr2027_conference,times}
\usepackage[utf8]{inputenc}
\usepackage[T1]{fontenc}
\usepackage{hyperref}
\usepackage{url}
\usepackage{booktabs}
\usepackage{amsfonts}
\usepackage{nicefrac}
\usepackage{microtype}
\usepackage{xcolor}
\usepackage{wrapfig}
\usepackage[font=footnotesize,labelfont=footnotesize]{caption}
\newcommand{\mypar}[1]{\par\noindent\textbf{#1}\ }
\usepackage{amsmath}
\newcommand{\bo}{\texttt{Boxoffice}}
\renewcommand{\paragraph}[1]{\mypar{#1}}

\usepackage{todonotes}

\usepackage{floatrow}
\usepackage{graphicx}
\usepackage{multirow,multicol}
\usepackage{tikz}
\usepackage{comment}
\usepackage{algpseudocode}
\usepackage{float}

\title{Evaluating the accuracy of KV cache reuse techniques}

\author{Samuel Cestola$^{1}$, Tianxiang Xia$^{2}$, Pengfei Zheng$^{1}$, Weiyan Zheng$^{1}$,\\
\textbf{Bo Wang$^{1}$, Yi Zhao$^{1}$, Diego Didona$^{1}$}\\
$^{1}$Huawei Technologies Ltd.\qquad $^{2}$Department of Computer Science, ETH Zurich}

\iclrfinalcopy
\begin{document}

\maketitle
\lhead{Preprint}
\vspace{-12pt}

\begin{abstract}
\looseness=-1 Position-independent KV cache reuse aims to reduce latency in retrieval-augmented generation by reusing chunk-level KV caches across prompts. We show that current evaluations of KV cache reuse techniques rely on measurements that fail to faithfully capture the loss of accuracy attributable to reuse, often artificially inflating the reported effectiveness. We also show that existing datasets do not exhibit the reuse dynamics needed to thoroughly evaluate such techniques. To address these issues, we propose an evaluation methodology that measures this accuracy loss without ambiguity and we introduce \texttt{Boxoffice}, a tool that programmatically generates evaluation datasets that exercise challenging KV cache reuse patterns.
\end{abstract}

\section{Introduction}
\looseness=-1 Retrieval-augmented generation (RAG) systems augment user queries with retrieved text chunks, supplying large language models (LLMs) with up-to-date and domain-specific context that improves answer accuracy and reduces hallucinations~\citep{lewis2021retrievalaugmentedgenerationknowledgeintensivenlp,metis,hydrarag}. The additional text chunks, however, inflate the prompt. Because attention cost is quadratic in prompt length, the resulting prefill stage, where the Key-Value (KV) caches are computed, becomes the dominant component of latency~\citep{transformer}. A promising direction for mitigating this overhead is position-independent KV cache reuse~\citep{cacheblend,epic,cachecraft,fusionrag}: each chunk's KV cache is computed once, stored, and subsequently reused at any position within future prompts that contain it. Position-independent KV cache reuse is strictly more general than prefix caching, the current industry standard~\citep{kwon2023efficientmemorymanagementlarge,zheng2024sglangefficientexecutionstructured}, which amortizes prefill cost only when chunks recur as the same ordered prefix. 

The main challenge in reusing KV caches at arbitrary positions is the absence of cross-chunk attention: tokens in one chunk carry no attention to tokens in other chunks. This is critical for accuracy: naive reuse of independently-prefilled chunks leads to severe accuracy degradation~\citep{cacheblend,gim2024promptcachemodularattention}. The widespread solution to recover cross-chunk attention is to recompute the KV cache for few tokens  at query time.  To preserve the performance benefits of KV cache reuse, such recomputation has to involve as few tokens as possible. An emerging line of work, which we refer to as {\em warm} approaches~\citep{cachecraft,fusionrag,lmcache}, aims to reduce the recomputation cost by embedding cross-chunk attention into the saved KV already at store time, so that less of it needs to be reconstructed at query time.%

\mypar{Pitfalls of KV cache reuse evaluations.} We identify  two issues in the evaluation methodology adopted in the literature on KV cache reuse approaches. %

\looseness=-1 The first issue is methodological. The headline metric for evaluating KV cache reuse should measure its core objective: preserving full prefill accuracy on queries where the correct answer genuinely depends on the provided context. However, existing work evaluates performance by reporting an aggregate accuracy metric, such as $F_1$ score, computed over every query in a dataset. This includes queries the baseline cannot answer even with full prefill (so there is no accuracy to preserve), queries the model can answer from parametric memory alone (so the KV cache is not what produces the answer), and yes/no or disjunctive questions prone to lucky guesses. Including such queries conflates accidental regularization brought by approximated attention states, parametric leakage, and chance correctness into the accuracy metric; as we show in Section~\ref{sec:limitations:methodology}, this systematically inflates the reported accuracy of reuse techniques, masking the true loss of accuracy attributable to the approximation.

The second issue lies in the chunk and query distributions of the datasets themselves. We identify two structural properties of commonly used datasets that systematically mislead, or even prevent, the evaluation of emerging warm approaches.  The first property is {\em clustered queries}. The chunks in the context of a query tend to be semantically very similar, because they concentrate on the same topic, and chunks are hardly reused in queries with a substantially different context. In a more general case the same chunk is included in different queries with different contexts: a chunk on Richard Feynman could appear in prompts together with chunks on other physicists, but also Nobel prize winners, or bongo players.  %
 We show in Section~\ref{sec:limitations:offline} that this clustering effect makes it overly easy to identify offline what cross-attention to bake into KV caches at construction time, just by relying on semantic proximity of chunks and without requiring any knowledge about the query and context distributions.  The second property is the limited presence of cross-query chunk reuse. A direct approach to warm KV cache reuse is to run a full prefill the first time a chunk is encountered, so that the saved KV carries the cross-attention from the preceding chunks present in that first query. We show in Section~\ref{sec:limitations:online} that in the most widely used datasets, queries are mostly composed of chunks that appear only once across the workload. 

\mypar{Contributions.}\looseness=-1 We make two contributions. First, in Section 4,  we present an evaluation methodology that isolates the accuracy loss attributable to KV cache reuse by removing its principal confounds. We define the {\em meaningful subset} of a benchmark as the set of queries on which the baseline answers correctly with full prefill, the model cannot answer without context, and the answer space is non-trivial. We
quantify the gap from the aggregate scores reported by prior
work:  on the Q\&A subset of the widely used LongBench benchmark~\citep{bai2024longbench}, up to $42\%$ of a reuse method's $F_1$ comes from queries that the full-prefill baseline cannot answer at all.
Second, in Section 5,  we propose a tool that generates a new dataset, which we call \bo{}, specifically tailored to evaluate KV cache reuse strategies. \bo{} is {\em meaningful by construction}: chunks are drawn from a fictional corpus, so the LLM parametric memory cannot help,  answers have a low chance of being guessed by chance, and the baseline with full prefill answers the query correctly. \bo{} also guarantees  cross-query chunk reuse, and exhibit
controlled, nuanced reuse patterns, including a new dynamic we
call \emph{stale cache}, where the same chunk is reused in
conflicting contexts. We show that the same reuse method can score an $F_1$ of $0.98$ or $0.00$ depending only on the context in which a chunk was cached in the first place. These dynamics reproduce across three query templates and on models of different sizes (8B and 32B), and the reuse regime \bo{} generates matches what production RAG traces report. %
 Code and datasets are available at~\citep{boxofficedata,boxofficerepo}.

\section{Background on Position-independent KV cache reuse}\label{sec:background}
A Transformer~\citep{transformer} LLM consists of stacked layers; within each layer, input
activations are linearly projected to query, key, and  value matrices $Q, K, V$, and self-attention is computed as
$A(Q, K, V) = \mathrm{Softmax}(QK^\top / \sqrt{d_k})\,V$. Inference proceeds in two phases:
\emph{prefill}, which processes the input prompt and computes
 $K$ and $V$, and \emph{decode}, which generates one
output token per step. Prefill cost is quadratic in prompt length; the
{\em KV cache} stores $K$ and $V$ so that decode reuses them. RAG systems prepend retrieved text chunks to the user query, at the cost of inflating the prompt and the prefill stage. Position-independent KV cache reuse, hereafter simply KV cache reuse, addresses this cost: each chunk's KV cache is computed once and can be reused at any position in other queries.

\begin{figure*}[t]
\centering
\captionsetup{skip=2pt}
\includegraphics[width=0.78\textwidth,trim={5 230 5 210},clip]{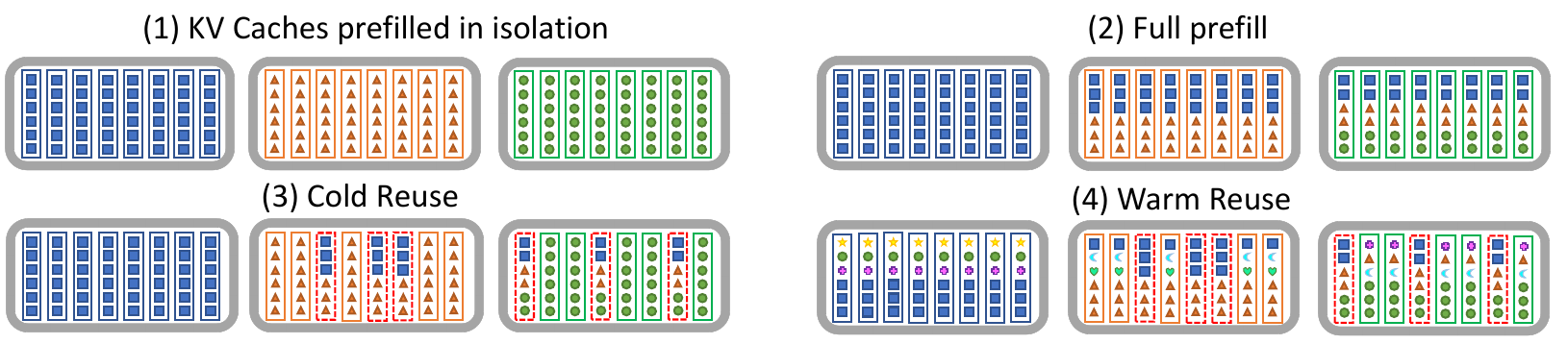}
\caption{KV cache reuse visualization for a query over $3$ chunks. Each gray box is the KV cache of
one chunk, the rectangles inside it are its tokens; each
chunk is associated with a shape;  the shapes inside a token denote the chunks towards
which it stores attention; dashed outlines mark tokens
whose attention is recomputed. \textbf{(1)}
{\em KV caches prefilled in isolation}: each cache stores
intra-chunk attention only. \textbf{(2)} \textit{Full
prefill}: each cache additionally stores cross-chunk attention
to other previous chunks in the prompt. \textbf{(3)} \textit{Cold
reuse}: start from~(1) and recover 
cross-chunk attention through selective recomputation.
\textbf{(4)} \textit{Warm reuse}: start from caches
that already store cross-chunk attention to other chunks, some
of which coincide with the prompt's chunks and some of which
do not (e.g.\ moon, stars); additionally recover the
cross-chunk attention to the prompt's chunks through
selective recomputation.}
\label{fig:reuse-regimes}
\end{figure*}

\mypar{Challenges of KV cache reuse.}
A reused KV cache differs from the one a normal prefill would compute over
the current prompt. First, the cached $K$ entries encode
the chunk's original positional embedding rather than its
position in the new prompt; for RoPE-based models~\citep{su2023roformerenhancedtransformerrotary} this is corrected by rotating $K$ to the new offset~\citep{cacheblend,lu2024turboragacceleratingretrievalaugmentedgeneration}. Second, the cached attention states encode no
\emph{cross-chunk} attention: tokens in one chunk carry no attention
to tokens in any other chunk, whereas under full prefill every token
attends to the preceding  tokens. 
The widespread solution to this issue is \emph{selective recomputation}, depicted in Figure \ref{fig:reuse-regimes}: at
query time, a fraction of tokens is selected and their attention
values are recomputed against the reused $K$ and $V$,
recovering part of the missing cross-attention~\citep{cacheblend}.

\mypar{Cold KV cache reuse.} The first breed of systems, which we refer to as {\em cold}, recover cross-attention only by means of selective recomputation, and differ in how they select the tokens to recompute.
CacheBlend~\citep{cacheblend} selects the tokens whose
freshly computed KV at layer 1 diverges the most from the cached values, i.e., the ones affected the most by the missing cross-chunk attention.
Following systems, such as Cacheclip~\citep{cacheclip} and ProphetKV~\citep{wang2026prophetkv}, identify, via cheap surrogates of the query-to-cache attention, the tokens the query will attend to most, and recompute that
subset. Across all cold variants, however, approximating full prefill accuracy requires
recomputing a substantial fraction of tokens at query time.

\mypar{Warm KV cache reuse.}
A second line of work, which we refer to as \emph{warm}, embeds cross-chunk attention into the saved KV \emph{at store time}, so that
less has to be recovered at query time.
FusionRAG~\citep{fusionrag} generates the KV cache of a chunk $C$ by prepending the chunk with its top-$k$ nearest chunks $T$ retrieved
from the corpus by embedding similarity, so that the KV cache of $C$ encodes the cross-chunk attention to $T$. Such {\em offline} approach assumes that if a query requires $C$, it also needs chunks similar to $C$.

\emph{Online} warm methods build the warm cache opportunistically
from the observed workload, so that the chunks that appear together in the prompts are determined by actual queries. LMCache~\citep{lmcache} runs a full prefill
the first time a chunk $C$ appears in a query, and saves the corresponding KV cache, which hence carries the cross-attention from the chunks preceding $C$ in that first query. CacheCraft~\citep{cachecraft} extends this idea with multi-versioning: it keeps several saved KVs per chunk, one for
each distinct prefix the chunk has been seen with, and at query time
selects the version whose prefix shares the most chunks with the
current prompt.
\section{Testing environment}\label{sec:testbed}
\mypar{Tasks.} KV cache reuse has been investigated across three main tasks:
question answering (QA), summarization, and agentic workloads. QA is
the most amenable to systematic study: {\em multi-hop} datasets come with the guarantee that providing the correct answer requires cross-attention among chunks,  
with a clear correct answer, and with an explicit set of \emph{golden chunks}
(the ones that actually contribute to the answer). %
 Summarization tasks exhibit
no reuse because chunks of a text to summarize appear only once in the dataset.  %
In agentic workloads, prompts differ only in specific portions~\citep{kvcomm,cacheslide}, or are the same but processed by different models~\citep{kvcomm,liu2025droidspeakkvcachesharing}; solutions tailored to them do not apply to the RAG case we study, where queries and contexts differ at arbitrary points.

\mypar{Datasets.} Q\&A is by far the most studied task in the KV cache reuse
literature. We find that the vast majority of papers evaluate on a common set of
three multi-hop subsets of LongBench~\citep{bai2024longbench}:
MuSiQue, HotpotQA, and 2WikiMQA. We
therefore focus our study on
these three datasets, which we refer to as {\em LongBench-QA}, that comprise 200 points each. Appendix~\ref{appendix:rw} 
breaks down the prevalence of each dataset across the surveyed
papers.

\mypar{Models and metrics.} For the core of our experiments we use LLama3.1-8B~\citep{grattafiori2024llama3herdmodels}, Qwen3-8B~\citep{yang2025qwen3technicalreport}, and Mistral-7B~\citep{jiang2023mistral7b}, which are used the most in existing experimental studies. We run a subset of the experiments of Sections 4 and 5 on Qwen3-32B to show that the dynamics that we present generalize to larger-size models.
 We use \texttt{e5-base-v2}~\citep{wang2022e5} to compute the embeddings of the chunks, and cosine similarity to measure similarity of embeddings. We use $F_1$ score as metric to evaluate accuracy~\citep{f1}, and we  report the  norm-$F_1$, i.e., the $F_1$ score obtained by a KV cache reuse method normalized w.r.t. that obtained with full prefill, so a value of $<1$ indicates loss of accuracy.

\mypar{KV cache reuse techniques.}\looseness=-1 We consider the KV cache reuse techniques implemented by the following state-of-the-art systems, described in Section 2. CacheBlend ($CB$, offline, cold), 
FusionRag ($FR$, offline, warm), LMCache ($LM$, online, warm), Cachecraft ($CC$, online, warm).  By default, tokens for recomputation are selected as in Cacheblend, by K matrix deviation at layer 1. We also implement a query-driven variant (noted $Q$) that selects the tokens as in FusionRag by identifying the highest-attention tokens via a lightweight computation between the last-layer query and K caches. %
 CacheCraft$-M$ runs a full prefill whenever it encounters a chunk with fewer than $M$ stored
warm caches; the case $M{=}1$ recovers \textit{LMCache}.  We implement these variants in a single framework based on HuggingFace Transformers~\citep{wolf2020huggingfacestransformersstateoftheartnatural}.
Unless stated otherwise, $R=15\%$ of the tokens are selectively recomputed to recover cross-attention, a common setting in experimental evaluations.

\section{Limitations of existing studies}\label{sec:limitations}
\begin{table*}
\scriptsize
\centering
\setlength{\tabcolsep}{3.5pt}
\resizebox{\textwidth}{!}{\begin{tabular}{c l rrr rrr rrr || rrr rrr rrr}
\toprule
 & & \multicolumn{9}{c}{$\rho_{F_1} = \mathrm{normF1}(\mathrm{filter}) / \mathrm{normF1}(\mathrm{ALL})$} & \multicolumn{9}{c}{ B0 ablation (200 points per dataset)} \\
\cmidrule(lr){3-11}\cmidrule(lr){12-20}
 & & \multicolumn{3}{c}{qwen} & \multicolumn{3}{c}{llama} & \multicolumn{3}{c}{mistral} & \multicolumn{3}{c}{qwen} & \multicolumn{3}{c}{llama} & \multicolumn{3}{c}{mistral} \\
\cmidrule(lr){3-5}\cmidrule(lr){6-8}\cmidrule(lr){9-11}\cmidrule(lr){12-14}\cmidrule(lr){15-17}\cmidrule(lr){18-20}
 & method / filter & $B_0$ & $+\text{NC}$ & $+\text{LI}$ & $B_0$ & $+\text{NC}$ & $+\text{LI}$ & $B_0$ & $+\text{NC}$ & $+\text{LI}$ & $B_0M_0$ & $B_0M_+$ & $\%F_1$ & $B_0M_0$ & $B_0M_+$ & $\%F_1$ & $B_0M_0$ & $B_0M_+$ & $\%F_1$ \\
\midrule
\multirow{4}{*}{\rotatebox[origin=c]{90}{musique}} & CB & 0.76 & 0.69 & 0.69 & 0.91 & 0.87 & 0.87 & 0.70 & 0.80 & 0.80 & 95 & 26 & 24.1 & 108 & 13 & 9.5 & 96 & 36 & 29.8 \\
 & CB+Q & 0.72 & 0.58 & 0.58 & 0.82 & 0.83 & 0.83 & 0.68 & 0.71 & 0.71 & 93 & 28 & 27.6 & 104 & 17 & 18.1 & 91 & 41 & 32.5 \\
 & FR & 0.74 & 0.64 & 0.64 & 0.83 & 0.70 & 0.70 & 0.58 & 0.67 & 0.67 & 98 & 23 & 26.4 & 103 & 18 & 16.9 & 121 & 11 & 42.4 \\
 & FR+Q & 0.73 & 0.65 & 0.65 & 0.75 & 0.70 & 0.70 & 0.63 & 0.79 & 0.79 & 95 & 26 & 26.6 & 97 & 24 & 25.4 & 123 & 9 & 36.8 \\
\midrule
\multirow{4}{*}{\rotatebox[origin=c]{90}{2wiki}} & CB & 0.87 & 0.58 & 0.52 & 0.88 & 0.84 & 0.86 & 0.96 & 0.81 & 0.79 & 89 & 15 & 13.2 & 73 & 13 & 12.5 & 61 & 12 & 4.5 \\
 & CB+Q & 0.75 & 0.49 & 0.45 & 0.88 & 0.73 & 0.73 & 0.94 & 0.75 & 0.62 & 83 & 21 & 24.9 & 72 & 14 & 12.3 & 61 & 12 & 5.8 \\
 & FR & 0.76 & 0.70 & 0.66 & 0.84 & 0.73 & 0.68 & 0.95 & 0.82 & 0.70 & 81 & 23 & 24.0 & 71 & 15 & 15.8 & 64 & 9 & 4.8 \\
 & FR+Q & 0.81 & 0.70 & 0.69 & 0.82 & 0.66 & 0.71 & 0.90 & 0.81 & 0.75 & 87 & 17 & 18.5 & 68 & 18 & 18.4 & 61 & 12 & 9.5 \\
\midrule
\multirow{4}{*}{\rotatebox[origin=c]{90}{hotpot}} & CB & 0.93 & 0.91 & 0.92 & 0.95 & 0.90 & 0.90 & 0.92 & 0.94 & 0.92 & 50 & 14 & 7.3 & 56 & 10 & 5.2 & 43 & 30 & 8.1 \\
 & CB+Q & 0.94 & 0.77 & 0.79 & 0.97 & 0.94 & 0.96 & 0.92 & 0.94 & 0.93 & 51 & 13 & 5.9 & 62 & 4 & 3.3 & 46 & 27 & 7.5 \\
 & FR & 0.94 & 0.82 & 0.80 & 0.93 & 0.80 & 0.81 & 0.93 & 1.01 & 0.98 & 54 & 10 & 5.9 & 55 & 11 & 6.5 & 58 & 15 & 6.6 \\
 & FR+Q & 0.97 & 0.90 & 0.89 & 0.94 & 0.85 & 0.84 & 0.93 & 0.97 & 0.94 & 56 & 8 & 3.0 & 55 & 11 & 5.6 & 59 & 14 & 7.1 \\
\midrule
\bottomrule
\end{tabular}}
\caption{\textbf{Left:} $\rho$ under the progressive filters of
Section~\ref{sec:limitations:methodology}; $\rho<1$ means measuring $F1$ on the unfiltered dataset inflates the reported accuracy. \textbf{Right:} $B0$ ablation.
$B_0M_0$ counts queries on which baseline and method both fail;
$B_0M_+$ counts queries on which the baseline fails but
the method succeeds; $\%F_1$ is the share of the unfiltered
aggregate $F_1$ contributed by $B_0M_+$.}
\label{tab:pitfalls}
\end{table*}

\subsection{Evaluation methodologies: pitfalls and our solution.}\label{sec:limitations:methodology}
\mypar{Current pitfalls.} To the best of our knowledge, all experimental analysis of KV cache reuse systems  report aggregate accuracy metrics, e.g., $F_1$, computed over every query of the evaluation set for the full prefill baseline and for the target system. 
We argue that this is the wrong evaluation methodology. The target metric for evaluating KV cache reuse should measure the ability to preserve full prefill accuracy on queries where the correct answer genuinely depends on the provided context. An aggregate over every query in a benchmark conflates this metric with three distinct phenomena: $i)$ accidental regularization brought by approximated attention states, which may arise on queries that the model fails to answer with full prefill; $ii)$  parametric leakage, which may arise on queries the model answers from parametric memory alone; and $iii)$ chance correctness, which may happen on yes/no or disjunctive questions. 

\mypar{Proposed solution.} We suggest to exclude three classes of queries from evaluation datasets: $i)$ queries where full prefill achieves 0 accuracy score with the target context. On these queries there is no accuracy-preservation capability to evaluate; any score gain achieved by the reuse strategy is due to accidental regularization brought by the approximated attention states; %
 $ii)$ queries that the model can answer correctly without additional context. The model can answer such queries without looking at the KV caches, so they are unfit to capture the ability of KV cache reuse strategies to preserve the model's ability to extract information from the context; and $iii)$ queries with a low-information answer (hereafter {\em low-info}), namely boolean yes/no queries and disjunctive queries of the form {\it who is richer, Elon Musk or Jeff Bezos?} The answer to such queries can be guessed correctly with probability 50\% by either the baseline or the reuse strategy.  Low-info queries may seem
addressable by prompt-based mitigations (``use only the provided
context''), but such instructions are not reliable~\citep{pezeshkpour2023mcq,zheng2024mcq,sclar2024prompt}. 

We refer to the queries that survive all three filters as the {\em meaningful subset} of the benchmark, since they are the ones on which the loss of accuracy attributable to KV cache reuse is observable.

\mypar{Impact of the evaluation methodology.} We empirically quantify the impact of each filter on LongBench-QA,
and show that reporting aggregate accuracy over all queries
systematically overstates how well a KV cache reuse method
preserves full prefill accuracy. We define three filters over the queries of a benchmark.
\textbf{B0} drops queries on which the full prefill baseline
scores $F_1 = 0$ on the target context. \textbf{NC} drops queries that the model can answer
without any context, identified as queries on which the
model achieves $F_1 > 0.2$ when prompted with the question
alone. \textbf{LI} drops yes/no and disjunctive queries. For each method, we measure the norm-F1 obtained on the
dataset that results from progressively applying these filters,
and report the ratio
$\rho = \text{norm-F1}_{\text{filtered}} / \text{norm-F1}_{\text{whole}}$;
$\rho < 1$ indicates that the filtered dataset reveals more
accuracy loss than the unfiltered headline reports. We compute
$\rho$ for \textbf{CB}, \textbf{CB+Q}, \textbf{FR}, and
\textbf{FR+Q}, which span cold (\textbf{CB}, \textbf{CB+Q}) and
warm-offline (\textbf{FR}, \textbf{FR+Q}) reuse, to show that
the conclusion is not specific to a single reuse strategy. We
exclude warm-online methods (LM, CC) because, as we will show
in Section~\ref{sec:limitations:online}, LongBench-QA does not exhibit the
cross-query chunk recurrence that those methods require. 

The results in Table~\ref{tab:pitfalls} (left) show
that $\rho$ generally decreases as we restrict the
evaluation dataset, indicating that each class of filtered-out
queries contributes to inflating the accuracy score aggregated
over the whole dataset. The B0 filter has the largest
individual impact, so we provide an additional ablation
(Table~\ref{tab:pitfalls}, right). For each (method,
dataset, model) cell we count the queries on which both the
baseline and the method score $F_1 = 0$ (\textbf{B0M0}), and
those on which the baseline scores $F_1 = 0$ but the method
scores $F_1 > 0$ (\textbf{B0M+}). The latter set is the
queries on which the approximate reused cache flips the answer
from totally wrong to at least partially right. We measure the
percentage of the aggregate $F_1$ score attributable to B0M+
queries; by construction it equals $1-\rho$ after B0, so B0 quantifies this inflation exactly, while NC and LI remove inflation not implied by construction.

\begin{table*}[t]
\centering
\scriptsize
\setlength{\tabcolsep}{3.5pt}
\begin{tabular}{l|c|c|c||c|c|c||c|c|c}
\toprule
 & \multicolumn{3}{c||}{qwen} & \multicolumn{3}{c||}{llama} & \multicolumn{3}{c}{mistral} \\
\cmidrule(lr){2-4}\cmidrule(lr){5-7}\cmidrule(lr){8-10}
method & 2wiki & hotpot & musique & 2wiki & hotpot & musique & 2wiki & hotpot & musique \\
\midrule
\textsc{CacheBlend (no infusion)}           & 0.13 & 0.21 & 0.10 & 0.38 & 0.35 & 0.19 & 0.14 & 0.19 & 0.13 \\
\textsc{Random}         & 0.39 & 0.52 & 0.35 & 0.48 & 0.55 & 0.43 & 0.32 & 0.48 & 0.27 \\
\textsc{Similarity-no-ctx} & 0.35 & 0.47 & 0.37 & 0.44 & 0.56 & 0.41 & 0.28 & 0.44 & 0.19 \\
\textsc{Context-nogold}   & 0.44 & 0.56 & 0.48 & 0.53 & \textbf{0.66} & 0.53 & 0.41 & \textbf{0.62} & 0.32 \\
\textsc{Golden-only}         & \textbf{0.63} & \textbf{0.67} & \textbf{0.52} & \textbf{0.57} & \textbf{0.66} & {\bf 0.56} & \textbf{0.52} & 0.59 & \textbf{0.34} \\
\textsc{FusionRag (similarity)}           & 0.55 & 0.59 & 0.36 & \textbf{0.57} & 0.61 & 0.42 & 0.42 & 0.55 & 0.20 \\
\bottomrule
\end{tabular}
\caption{\looseness=-1 Mean norm-F1 of different offline cross-chunk attention infusion modes. The provided cross-chunk attention helps significantly over a random baseline only if it aligns with the query's context or golden chunks.}
\label{tab:infusion}
\end{table*}

The data shows that, first, a significant fraction of queries
are unanswerable by either the baseline or the method,
consistent with the low aggregate $F_1$ scores typically
reported on LongBench-QA (between $0.2$ and $0.5$). Second,
flipping queries contribute substantially to the aggregate
$F_1$ score, reaching $42\%$ on MuSiQue and $24.9\%$ on
2WikiMQA, the two harder datasets. This reinforces the message
that including baseline-zero queries inflates the $F_1$ scores
reported by existing studies through dynamics that are
orthogonal to the preservation of full prefill accuracy. 

Appendix~\ref{app:tau} shows that the results presented hold for any threshold in $[0,0.5]$, reports the size of the meaningful subset per dataset and model, and discusses its model dependence. Appendix~\ref{app:32b} shows that the $F_1$ inflation phenomenon persists, attenuated, on a $4\times$ larger model (Qwen3-32B).

\looseness=-1 \textbf{Discussion.} The classes of queries we filter out may 
capture useful side-effects of KV cache reuse, e.g., accidental regularization on
long contexts, as documented in sparse attention and KV cache compression~\citep{xiao2024efficientstreaminglanguagemodels} evaluation studies. In those cases, however, the  effect of such regularization  is infrequent and mild, which may justify an aggregate accuracy score over an unfiltered dataset. In our study, instead, such accidental regularization significantly impacts accuracy and shows up consistently across methods, models, and datasets. In the context of reuse, these phenomena deserve their own investigation, but existing evaluations provide no ablation studies in this direction. Our proposed methodology measures the accuracy degradation attributable to KV cache reuse in isolation.

\subsection{Dataset limitations for warm reuse - the offline case}\label{sec:limitations:offline}

\mypar{What cross-attention is needed?} We run an experiment to evaluate the effect of injecting different cross-chunk attention to KV caches used to serve a query. For each chunk in a query, we construct its KV cache with a
prefix of additional chunks; the prefix determines what
cross-attention the cache stores. We test six prefix
conditions per chunk in the context $T$: no prefix (plain
\textit{CacheBlend}); random chunks
drawn from the dataset; semantically close chunks that are
not in $T$; the golden chunks of $T$; non-golden chunks that
appear in $T$; and ten semantically close chunks, as in
\textit{FusionRAG}.  To isolate the effect
of cross-attention infusion we set $R=0$.
Table~\ref{tab:infusion} reports the resulting norm-F1 on the meaningful subset of queries identified in the previous section.
Cross-attention infusion improves over plain Cacheblend, but
only when the injected attention is drawn from chunks that
actually appear in the target query, with golden chunks
providing the largest gain. Cross-attention from semantically
close chunks that do \emph{not} appear in $T$ offers no consistent benefit
over random chunks. The random variant itself is already
better than the no-infusion case, which we
attribute to infusion suppressing the attention sinks that distort
the dynamics of independently-prefilled KV
caches~\citep{xiao2024efficientstreaminglanguagemodels,ape,epic}.

\begin{table}[t]
\centering
\footnotesize
\setlength{\tabcolsep}{3pt}
\caption{\footnotesize \textbf{Left:} $CC$ ($GC$) is the percentage of a query's
context (golden) chunks identifiable by cosine similarity alone,
without knowledge of the query. High values indicate that warm
offline retrieval can identify the right cross-attention targets
purely from chunk embeddings. \textbf{Middle:} mean cosine
similarity between chunk pairs $g$: goldens of the query; $c$: non-golden context chunks
of the query; $o$: chunks outside the query. Golden chunks and context chunks are clustered and easily
separable from chunks of other queries. \textbf{Right:} reuse
statistics. $C$: \% of unique chunks that appear in $\geq 2$
queries. $G$: among the reused chunks, the \% that are golden in
some query. $F$: \% of queries whose entire context has already
appeared in earlier queries.}
\label{tab:warm}
\begin{tabular}{l cc || cccccc || ccc}
\toprule
Dataset & $CC$ & $GC$ & $g \to g$ & $g \to c$ & $g \to o$ & $c \to g$ & $c \to c$ & $c \to o$ & $C$ & $G$ & $F$ \\
\midrule
MuSiQue  & 0.54 & 0.69 & 0.77 & 0.73 & 0.70 & 0.73 & 0.74 & 0.70 & $12.6\%$ & $26.6\%$ & $0\%$ \\
2WikiMQA & 0.56 & 0.81 & 0.78 & 0.75 & 0.71 & 0.75 & 0.75 & 0.70 & $11.9\%$ & $\phantom{0}0.7\%$ & $0\%$ \\
HotpotQA & 0.84 & 0.88 & 0.81 & 0.78 & 0.70 & 0.78 & 0.78 & 0.70 & $\phantom{0}0.7\%$ & $16.7\%$ & $0\%$ \\
\bottomrule
\end{tabular}
\end{table}

\mypar{Important cross-attention is easily identified.}
On LongBench-QA, identifying golden and context chunks offline
is easy: the cosine-similarity neighborhood of
a chunk $C$ already contains most of the  context of the queries in which $C$ is used.
For a query $Q$ with context $T_Q$ and golden subset
$G_Q \subseteq T_Q$, let $\textsc{topk}(c)$ denote the $k$
nearest neighbors of $c$ in the corpus by cosine similarity,
and $U_k(Q) = \bigcup_{c \in T_Q} \textsc{topk}(c)$ the union of
these neighborhoods. We define \emph{context coverage}
$CC(Q) = |U_k(Q) \cap T_Q|/|T_Q|$ and \emph{golden coverage}
$GC(Q) = |U_k(Q) \cap G_Q|/|G_Q|$.
Table~\ref{tab:warm} (left) reports both at $k{=}10$:
offline neighbor-based infusion already provides cross-chunk
attention, on average, to at least $54\%$ of a query's context chunks and
$69\%$ of its golden chunks. Warm offline systems can therefore identify the right
cross-attention targets without any knowledge of the query.
The reason is a tight cluster structure: for each chunk in a query, we compute its cosine similarity
to every other golden chunk of the same query, every
non-golden context chunk of the same query, and every chunk
outside the query. We then report the average broken down by
the role of the two chunks in the pair: $g$ (golden), $c$
(non-golden context), $o$ (outside the query).
Table~\ref{tab:warm} (right) shows that golden chunks lie close to other golden chunks, and context chunks lie close to other context chunks, especially the golden ones. Appendix~\ref{appendix:cdf} visualizes this through the distance CDFs.
This tight clustering is not representative of a broad class of
RAG workloads where the same chunk participates in queries
that involve different contexts. Our analysis of industry RAG workloads (Appendix~\ref{app:traces}) shows that individual chunks recur often across queries, but the chunks they appear with rarely do: two
  requests that share a chunk have a median Jaccard similarity of only about 0.1 between their chunk sets, and a chunk's actual co-used partner is among
  its 10 nearest neighbours computed offline in only 8\% of cases. These workloads challenge the core assumption behind chunk-similarity-based approaches.

\subsection{Dataset limitations for warm reuse - the online case}\label{sec:limitations:online}
\looseness=-1 We analyse whether LongBench-QA is fit to evaluate warm online KV cache reuse
methods. The main required property is that the same chunk
appears in at least two queries: one occurrence seeds the warm KV
cache, the other is the query on which reuse is evaluated;
without it, no KV cache reuse happens.

We further identify two desirable properties.  First, there must exist {\em queries whose chunks are  all reused}. This enables steady-state evaluation of warm systems: on a chunk's first appearance, a warm system may run a partial or full prefill~\citep{lmcache, cacheblend, kvcomm}, hiding the accuracy degradation that would arise at steady state under full reuse. Full reuse is the regime that the evaluations of all cold methods reproduce, and it is also the common case in practice: in the RAGPulse production trace, 73\% of first-week requests (93\% after one week of warmup) consist entirely of previously seen chunks, as do 40--50\% of the questions in the HERB enterprise benchmark.
Second, {\em the reused chunks should be golden}. The purpose of reuse is to preserve/restore, the attention states that
matter for answering the query. If reuse events are concentrated on
{\em distractor} chunks, the evaluation cannot evaluate if  a method can preserve/restore the meaningful cross-attention.

\noindent\textbf{Existing datasets are unfit for warm KV cache reuse.} %
 Table~\ref{tab:warm} (right) reports three  metrics computed on LongBench-QA:
 $i)$ {\em chunk reuse} ($C$), i.e.,  the percentage of unique chunks
that appear in two or more queries; $ii)$ {\em golden reuse} ($G$), i.e., the percentage of golden chunks among the reused chunks; and $iii)$ {\em full-context reuse} ($F$), i.e., the fraction of queries whose chunks have  {\em all} appeared already in previous queries. $C$ lies between $0.7\%$ and $12.6\%$, and re-used chunks are golden chunks only in a minority of the cases (between $0.7\%$ and $26.6\%$). Finally, no
query can be served entirely from re-used KV caches, meaning that warm online methods would run at
least a partial prefill on every query of the dataset to populate the KV cache.

\section{Boxoffice: a stress-test benchmark for KV cache reuse}
\label{sec:boxoffice}
\bo{} is a tool that programmatically generates
a synthetic dataset designed  to evaluate KV cache reuse techniques. \bo{} produces a corpus of
structured records describing fake movies, and a set of queries on
such corpus. Each record describes a fictional movie with a small set of
fields, prominently a numeric box office field, plus short 
paragraphs that paraphrase the schema content. Each query asks to
identify the movie with the highest box office
value among a fixed-size candidate set provided in the context (whose default size is 10). This argmax template is the default one, which we discuss in detail in this Section. Appendix~\ref{app:templates} presents two additional templates, a categorical uniqueness task and a two-hop pivot join, and shows that the dynamics we present generalize to other query templates. 
\looseness=-1 \bo{} addresses all the limitations that we have discussed so far, and is intended as a controlled stress-test complement to existing benchmarks rather than as a replacement.

\mypar{Meaningful by construction.}
Because the films are fictional, queries cannot be answered from parametric memory of the model. Moreover, \bo{} includes in the produced
datasets only queries that can be answered correctly with full
prefill by the target model(s). Finally, with 10 candidates per
query, the chance that a randomly guessed answer is correct is low (0.1). %

\mypar{Heterogeneous contexts.}
Queries draw from a shared pool of chunks, and the same chunk
appears alongside different context chunks in different queries (on the released datasets, every candidate slot holds a reused chunk and the median Jaccard similarity between queries sharing a chunk is $0.11$, as in the workloads of Section~\ref{sec:limitations:offline}).
The within-query semantic clustering identified in
Section~\ref{sec:limitations} cannot occur on \bo{}. For a corpus of
$100$ movies, $k{=}5$ and $10$ chunks per query, golden coverage
at $k{=}5$ using chunk semantic similarity is $0.32$, no better than the $0.37$ that random
sampling of $5$ chunks from the $100$-movie corpus would yield in
expectation. The mean pairwise cosine similarity is $0.89$ both
within a query and from a golden chunk to chunks outside the query
(see also Appendix~\ref{appendix:cdf}): query membership carries no embedding signal, so top-$k$
similarity cannot distinguish in-query from outside-query chunks.

\mypar{Programmatic golden reuse.}
\bo{} splits queries into a \emph{warmup} set and an
\emph{evaluation} set: every chunk used
in the evaluation set has already appeared in at least one warmup
query, so all evaluation queries rely on chunks with a
re-usable KV cache. A configurable parameter $M$
controls the minimum number of warmup appearances per evaluation
chunk. This feature enables the evaluation of systems that maintain
$M$ different KV caches of the same chunk, such as CacheCraft. Note
that in \bo{} every chunk is effectively golden: the answer is the
argmax of the box office field across the candidate set, so every
candidate dossier contributes to answering correctly.
\subsection{Cache staleness, a new KV cache reuse dynamic}
\bo{} controls not only whether a chunk has been seen before but
also the role under which it was seen. For any query, we
call the \emph{winner} the chunk whose box office value is maximal
among the context candidates and the \emph{nonwinners} the remaining
ones. During warmup, a target chunk is cached after predecessor
chunks whose box office values are either all strictly smaller than
the target's, in which case the target was the winner of its
warmup query, or all strictly larger, in which case the target was
a nonwinner. The cached state therefore carries a directional
signal: a winner-cached chunk implicitly encodes ``I am the maximum
among my predecessors'', and a nonwinner-cached chunk encodes the
opposite. We refer to a cached chunk as \emph{stale} when the role
it carries from warmup does not match its role in the query at
hand.

\mypar{Four warmup configurations.}\looseness=-1
Every evaluation query falls into one of four
warmup configurations, defined by the role under which each chunk
was cached when it appeared for the \emph{first} time in the
warmup. Two binary axes determine the configuration: whether the
evaluation winner was cached as a winner ($W$) or as a nonwinner ($N$),
and whether all evaluation nonwinners were cached as winners
 or as nonwinners. We label the four configurations
\texttt{aligned}, \texttt{all-W}, \texttt{all-N}, and
\texttt{flipped}; Table~\ref{tab:cells} describes each one,
the cached signals it presents to the reuse method, and the
qualitative difficulty it imposes. The evaluation set is balanced across
configurations.

\mypar{Staleness with multiple cached versions per chunk.}
When the warmup is configured with $M{>}1$ cached versions
per chunk, the first cached version reflects the
role the chunk plays at its first warmup appearance, and the
following $M{-}1$ appearances balance the role
distribution: by the end of warmup, at least
$\lfloor M/2 \rfloor$ appearances are in the
winner role and at least $\lceil M/2 \rceil$ in the
nonwinner role, or vice versa. Each chunk therefore has at least one
warmup appearance whose role matches its role in an
evaluation query and at least one whose role does not (e.g., with $M{=}2$, a chunk first cached as a winner is
later placed as a nonwinner). This construction evaluates whether a
method that stores $M{>}1$ versions per chunk, such as
\textit{CacheCraft}, can pick the version whose cached
signals best match the current query, and hence recover from
staleness that a single-version method can only address by
selective recomputation.

\subsection{Evaluating KV cache reuse with \bo{}}\label{sec:boxoffice:results}
We evaluate the full set of KV cache reuse methods introduced in Section~\ref{sec:background} on \bo{}. 
Table~\ref{tab:bo-15} reports norm-F1 (which equals $F_1$ by construction), providing the average value across all the evaluation queries, and a breakdown by configuration. The evaluation set includes 160 queries (40 per configuration); warmup queries are roughly 250 and guarantee that each chunk in the evaluation set appears at least 4 times each during warmup, to enable the evaluation of $CC-M4$. We average the results across 10 different generated datasets; sample standard deviations across seeds and results with $R=0.05$ are reported in Appendix~\ref{appendix:bo}.

\begin{table}[t]
\centering
\scriptsize
\renewcommand{\arraystretch}{0.95}
\setlength{\tabcolsep}{1pt}
\caption{norm-F1 on \bo{}. Within each (model, configuration), the winning
method is in bold.}
\label{tab:bo-15}
\begin{tabular}{l|*{10}{c} @{\hspace{10pt}} *{10}{c} @{\hspace{10pt}} *{10}{c}}
\toprule
 & \multicolumn{10}{c}{\textbf{Llama}} & \multicolumn{10}{c}{\textbf{Qwen}} & \multicolumn{10}{c}{\textbf{Mistral}} \\
\cmidrule(lr){2-11} \cmidrule(lr){12-21} \cmidrule(lr){22-31}
Config
& \rotatebox{90}{\,CB} & \rotatebox{90}{\,CB+Q}
& \rotatebox{90}{\,FR} & \rotatebox{90}{\,FR+Q}
& \rotatebox{90}{\,LM} & \rotatebox{90}{\,LM+Q}
& \rotatebox{90}{\,CC-M2} & \rotatebox{90}{\,CC-M2+Q}
& \rotatebox{90}{\,CC-M4} & \rotatebox{90}{\,CC-M4+Q}
& \rotatebox{90}{\,CB} & \rotatebox{90}{\,CB+Q}
& \rotatebox{90}{\,FR} & \rotatebox{90}{\,FR+Q}
& \rotatebox{90}{\,LM} & \rotatebox{90}{\,LM+Q}
& \rotatebox{90}{\,CC-M2} & \rotatebox{90}{\,CC-M2+Q}
& \rotatebox{90}{\,CC-M4} & \rotatebox{90}{\,CC-M4+Q}
& \rotatebox{90}{\,CB} & \rotatebox{90}{\,CB+Q}
& \rotatebox{90}{\,FR} & \rotatebox{90}{\,FR+Q}
& \rotatebox{90}{\,LM} & \rotatebox{90}{\,LM+Q}
& \rotatebox{90}{\,CC-M2} & \rotatebox{90}{\,CC-M2+Q}
& \rotatebox{90}{\,CC-M4} & \rotatebox{90}{\,CC-M4+Q} \\
\midrule
aligned & .10 & .31 & .11 & .37 & \textbf{.98} & .95 & .17 & .55 & .19 & .51 & .39 & .09 & .44 & .40 & \textbf{.97} & .94 & .17 & .38 & .18 & .25 & .20 & .18 & .33 & .53 & \textbf{.98} & .94 & .14 & .42 & .21 & .36 \\
all-W   & .06 & .21 & .05 & .28 & .14 & .29 & .13 & .42 & .24 & \textbf{.50} & .30 & .12 & .28 & .29 & .11 & .25 & .13 & .27 & .24 & \textbf{.32} & .24 & .15 & .15 & .26 & .17 & .26 & .14 & .26 & .31 & \textbf{.34} \\
all-N   & .06 & .25 & .13 & .43 & .07 & .55 & .16 & .57 & .23 & \textbf{.57} & .39 & .20 & .30 & \textbf{.39} & .05 & .11 & .18 & .36 & .24 & .34 & .29 & .16 & .24 & \textbf{.57} & .09 & .29 & .16 & .34 & .32 & .39 \\
flipped & .09 & .25 & .06 & .35 & .00 & .09 & .11 & .41 & .23 & \textbf{.54} & \textbf{.41} & .18 & .23 & .28 & .00 & .06 & .15 & .29 & .28 & .34 & .25 & .11 & .18 & \textbf{.34} & .00 & .02 & .13 & .23 & .30 & .30 \\
\midrule
average & .08 & .26 & .09 & .36 & .30 & .47 & .14 & .49 & .22 & \textbf{.53} & \textbf{.37} & .15 & .31 & .34 & .28 & .34 & .16 & .33 & .24 & .31 & .24 & .15 & .23 & \textbf{.42} & .31 & .38 & .14 & .31 & .28 & .35 \\
\textit{std} & .03 & .04 & .02 & .03 & .03 & .03 & .05 & .07 & .04 & .06 & .05 & .03 & .04 & .04 & .02 & .02 & .05 & .05 & .04 & .04 & .03 & .04 & .03 & .04 & .02 & .04 & .05 & .06 & .07 & .05 \\
\bottomrule
\end{tabular}
\end{table}

\mypar{Result 1. \bo{}'s warmup elicits configuration-dependent
reuse dynamics.}
\bo{}'s configuration-driven warmup biases each chunk's
cached KV strongly enough to drive the reuse dynamics visible
in Table~\ref{tab:bo-15}.  The clearest evidence is in the two extreme configurations on $LM$:
on \texttt{aligned}, where every cached KV encodes the same
role the chunk plays at query time, \textit{LM} achieves
near-perfect norm-F1 across all three models ($\geq 0.97$);
on \texttt{flipped}, where every cached role contradicts the
role the chunk plays at query time, \textit{LM} drops to
$0.00$ on every model. The intermediate configurations  sit between the two extremes:
the cached cross-attention is partially incorrect, and the
recomputation budget can only partially repair it. The same collapse and recovery appear on the two additional templates (Appendix~\ref{app:templates}) and with Qwen3-32B (Appendix~\ref{app:32b}).

\mypar{Result 2. Chunk similarity alone is not effective for warm
offline KV cache reuse on \bo{}.}\looseness=-1
\textit{FR} is on average on par with \textit{CB} on LLama and Mistral, and worse than it on Qwen. 
This is a direct consequence of \bo{}'s non-clustered chunk distribution
(Section~\ref{sec:boxoffice}). The Q-augmented variant \textit{FR+Q} outperforms
\textit{CB+Q} across all three models on average; we attribute this to the
warmup itself providing better KV cache conditioning (e.g., removing attention sinks~\citep{epic,ape}) that
\texttt{+Q} can leverage at recomputation time, not to
\textit{FR}'s neighbour-based mechanism: the same
dynamic arises under random offline warmup
(Section~\ref{sec:limitations:offline}).

\mypar{Result 3. Query-driven token selection helps the warm methods.}
On average, the \texttt{+Q} variant improves over the
deviation-based default on every warm method (\textit{FR},
\textit{LM}, \textit{CC-M2}, \textit{CC-M4}) and every model. The
picture is different for the cold method \textit{CB}: \texttt{+Q}
helps on Llama  but regresses on Qwen  and
Mistral, plausibly because cold KV caches encode
distorted attention patterns such as attention sinks~\citep{xiao2024efficientstreaminglanguagemodels} that confound a selection mechanism relying on attention
scores against the cached KVs. Even on the warm methods, \texttt{+Q} does not close the gap with
the full prefill baseline, although \bo{} makes the key
information for each query identifiable: better token-selection strategies are needed, and \bo{} provides a controlled testbed to develop them.

\mypar{Result 4. Multi-version KV caching is powerful but its
selection criterion matters.}
Multi-version methods reach the
highest overall norm-F1 on Llama, so storing multiple cached
versions per chunk pays off when the right
version is selected at query time. However, on \texttt{aligned}
queries, \textit{LM} ($M{=}1$) outperforms every $M{>}1$
configuration, because with
$M{>}1$ the system can pick the wrong cached version. \textit{CC}
selects a cache by prefix similarity, a criterion
oblivious to whether the chosen cache is aligned with the
role the chunk plays in the current query; with $10$ chunks
per query, even a small per-chunk probability of
mis-selection compounds, so at least one
chunk's cache is likely to break the perfect alignment.

\mypar{Result 5. No single strategy is robust across models.}\looseness=-1
On average, the winner differs on every model: \textit{CC-M4+Q} on Llama , \textit{CB} on Qwen, and \textit{FR+Q} on Mistral. These three winners
are heterogeneous in the hot/cold and online/offline taxonomy, so no family of KV cache
reuse strategies is preferable across models, and conclusions drawn from a single model
risk misgeneralising.

\section{Related work}\label{sec:rw}

\mypar{KV cache reuse methodologies and datasets.}
Existing evaluations of KV cache reuse~\citep{cacheblend,
fusionrag, cachecraft, lmcache} report aggregate accuracy on
standard QA benchmarks, primarily from
LongBench~\citep{bai2024longbench} (as shown in Appendix~\ref{appendix:rw}). We show in
Section~\ref{sec:limitations:methodology} that this inflates reported
retention, substantially
overstating the effectiveness of the techniques. To simulate
the cross-query reuse that standard QA datasets lack, two
simple workarounds have been used: LLM-generated questions
over a fixed context~\citep{cacheblend, feng2025evicpress}, and
partitioning a single long document into smaller chunks
queried by multiple questions~\citep{contextpilot}. Both have
limitations that \bo{} addresses. They yield clustered
contexts, since paraphrased questions re-target the same
chunks and chunks split from one document share a topic by
construction; they do not enforce golden-chunk reuse; and they
do not generate queries composed entirely of reused chunks.
The evaluation of CacheCraft~\citep{cachecraft} relies on a
proprietary workload that is not released. \bo{} enables the
reproducible evaluation of multi-versioned KV cache reuse
strategies such as CacheCraft, enforcing parametric reuse of
chunks so that enough versions accumulate.

\mypar{Evaluation beyond KV cache reuse.}\looseness=-1
Our methodology aligns with prior evidence that aggregate accuracy can be inflated by queries answered from dataset artifacts or memorised test items~\citep{gururangan2018annotation,mccoy2019right,sainz2023nlp,magar2022data}, with calls for per-capability rather than single-score evaluation~\citep{hooker2020characterising,ribeiro2020checklist,chen2024rgb}, and with work that separates the contribution of the context from that of parametric memory~\citep{mallen2023whennot,tao2024whencontext,longpre2021entitybased,xiong2024mirage}; synthetic or freshly sourced data has likewise been used to prevent parametric leakage~\citep{zhang2024infinitebench,white2024livebench,eyzaguirre2024cartridges,wei2023sycophancy}. See Appendix~\ref{app:rw-extended} for details.

These results and insights back the methodology we
propose to clearly measure the accuracy loss attributable to
KV cache reuse. \bo{} follows this methodology by construction, and
enables the evaluation of KV cache reuse strategies on
challenging cross-query reuse dynamics.
\section{Conclusions}\label{sec:conclusions}
We show limitations in the methodology and the datasets
currently used to evaluate KV cache reuse strategies, propose remedies for both, and release \bo{}, a tool that enables
the evaluation of KV cache reuse under challenging and
controllable reuse patterns, towards a more thorough and reproducible evaluation of KV cache
reuse strategies.
\mypar{Limitations.}\looseness=-1 First, \bo{} is built around three query templates over one synthetic corpus (Appendix~\ref{app:templates}), whereas
suites such as LongBench span multiple tasks. Second, our main evaluation covers three open-source $\sim 8$B
models; Appendix~\ref{app:32b} verifies on Qwen3-32B that the accuracy inflation and the staleness dynamic persist, but other architectures may
be affected quantitatively differently by the filtered-out queries, even though the methodological
argument is unchanged.
Third, we focus on recomputation-based KV
cache reuse, the family currently seeing industry
adoption~\citep{lmcache,cachecraft,ucm,wang2025mepic}. The inflation analysis and the meaningful subset apply unchanged to approaches based on fine-tuning or post-training~\citep{lu2024turboragacceleratingretrievalaugmentedgeneration,ma2025blockattentionefficientprefilling,yang2025kvlinkacceleratinglargelanguage}, since the filters only look at the answers a method produces; the staleness dimension is instead specific to methods that store context-conditioned caches, for which \bo{} serves as a leakage-free testbed.

\clearpage
\subsection*{AI use statement}
In this work, we used generative AI tools for editing (grammar, spelling, word choice), for retrieval and discovery of related work, and for assisting with code development (helper scripts for data processing, filtering, and figure generation); all AI-assisted code was reviewed and tested by the authors, and all results were produced by running the released code. We have not used generative AI tools to generate synthetic datasets (the \bo{} corpus and queries are produced programmatically by the released generator), to develop theoretical models, to propose hypotheses, to design experiments, or to interpret results. We take responsibility for the final content of this work, including text, claims, and artifacts produced with the aid of generative AI.

\subsection*{Reproducibility statement}
The code for every method, the LongBench-QA enrichment scripts, and the \bo{} generator (including the two additional templates of Appendix~\ref{app:templates}) are publicly available in a repository, \url{https://github.com/boxoffice1280/boxoffice1280}, and the exact generated datasets used in this paper in a dataset release, \url{https://huggingface.co/datasets/Boxoffice1280/boxoffice1280}~\citep{boxofficedata,boxofficerepo}. Section~\ref{sec:testbed} lists models, embedding model, metric, and recomputation budget; Appendix~\ref{app:bo-generation} gives the exact prompts and the generation pseudocode; Appendix~\ref{appendix:bo} reports per-seed variability and the runtime of an end-to-end run.

\bibliographystyle{iclr2027_conference}
\bibliography{nips}

\clearpage
\appendix
\section{Appendix - Prevalence of LongBench Datasets}\label{appendix:rw}
Table~\ref{tab:appendix-longbench-multidoc} shows which LongBench-QA datasets are used in the evaluation of state-of-the-art papers that investigate KV cache reuse. We find that such datasets are used each in more than $80\%$ of the papers we reviewed. 
Other datasets that are used with less frequency are TriviaQA~\citep{triviaqa}, Qasper~\citep{qasper}, NarrativeQA~\citep{narrativeqa}, Ruler~\citep{ruler}, MultihopRag~\citep{multihoprag}. These datasets present each some nondesirable property for evaluating KV cache reuse in the RAG case, e.g., being single-hop, or having a single long chunk with  multiple (similar and clustered) queries over it, or having only very few (e.g., 3) long chunks as context. As said in Section~\ref{sec:background} we do not consider summarization tasks because there is no reuse in those datasets, and agentic workload datasets because they target different re-use dynamics.

\begin{table*}[t]
\centering
\scriptsize
\setlength{\tabcolsep}{6pt}
\renewcommand{\arraystretch}{1.1}
\begin{tabular}{lccc}
\toprule
Paper & HotpotQA & 2WikiMQA & MuSiQue \\
\midrule
CacheBlend~\citep{cacheblend} & \checkmark & \checkmark & \checkmark \\
Cache-Craft~\citep{cachecraft} & $\times$ & \checkmark & \checkmark \\
CacheClip~\citep{cacheclip} & \checkmark & \checkmark & \checkmark \\
EPIC~\citep{epic} & \checkmark & \checkmark & \checkmark \\
SamKV~\citep{SamKV} & \checkmark & \checkmark & \checkmark \\
ContextPilot~\citep{contextpilot} & $\times$ & $\times$ & $\times$ \\
FusionRAG~\citep{fusionrag} & \checkmark & \checkmark & \checkmark \\
Droidspeak~\citep{liu2025droidspeakkvcachesharing} & \checkmark & \checkmark & \checkmark \\
ProphetKV~\citep{wang2026prophetkv} & \checkmark & \checkmark & \checkmark \\
CacheSlide~\citep{cacheslide} & \checkmark & $\times$ & $\times$ \\
KVLink~\citep{yang2025kvlinkacceleratinglargelanguage} & \checkmark & \checkmark & \checkmark \\
TurboRAG~\citep{lu2024turboragacceleratingretrievalaugmentedgeneration} & \checkmark & \checkmark & \checkmark \\
BlockAttention~\citep{ma2025blockattentionefficientprefilling} & \checkmark & \checkmark & \checkmark \\
APE~\citep{ape} & \checkmark & \checkmark & \checkmark \\
CacheGen~\citep{cachegen} & \checkmark & \checkmark & \checkmark \\
LMCache~\citep{lmcache} & \checkmark & \checkmark & \checkmark \\
A$^3$~\citep{zhou2025a3} & \checkmark & \checkmark & \checkmark \\
COMB~\citep{zhao2026comb} & \checkmark & \checkmark & \checkmark \\
EVICPRESS~\citep{feng2025evicpress} & \checkmark & \checkmark & \checkmark \\
InfoFlow KV~\citep{teng2026infoflow} & \checkmark & \checkmark & \checkmark \\
SEL~\citep{zhang2025attention} & \checkmark & $\times$ & $\times$ \\
MEPIC~\citep{wang2025mepic} & $\times$ & $\times$ & $\times$ \\
\citet{cestola2026experimental} & $\times$ & \checkmark & \checkmark \\
\midrule
Papers using benchmark & 19/23 & 19/23 & 19/23 \\
\bottomrule
\end{tabular}
\caption{Use of LongBench-QA datasets in state-of-the-art papers on KV cache reuse.}
\label{tab:appendix-longbench-multidoc}
\end{table*}
\section{Extended related work}\label{app:rw-extended}
This appendix expands the second part of Section~\ref{sec:rw}.

A line of work counters dataset contamination through
synthetic data or fresh sourcing: InfiniteBench~\citep{zhang2024infinitebench} replaces
core entities in real books with fictional ones, and
LiveBench~\citep{white2024livebench} sources fresh questions
over time, both to prevent answers from leaking via parametric
memory. Synthetic-corpus generation has also been used for
adjacent purposes, e.g.,  training compact long-context
representations~\citep{eyzaguirre2024cartridges} or surfacing
specific behaviors such as sycophancy~\citep{wei2023sycophancy}.

Work on natural language inference shows that aggregate
accuracy can be inflated by queries the model answers from
dataset artifacts rather than from actual reasoning about the
query~\citep{gururangan2018annotation, mccoy2019right}. Work on
benchmark contamination~\citep{sainz2023nlp, magar2022data}
documents the analogous effect of test items leaking into
pretraining, where aggregate scores are inflated by
memorisation. \citet{hooker2020characterising}
show how a subset of queries can disproportionately affect an
aggregate metric when evaluating  compressed models. \citet{ribeiro2020checklist}
make the same case for natural language processing evaluation more broadly, arguing
that a single aggregate score collapses distinct capabilities
and proposing per-capability tests. In the RAG setting,
RGB~\citep{chen2024rgb} stratifies accuracy into noise
robustness, negative rejection, information integration, and
counterfactual robustness, rather than a
single aggregate.

Prior work has measured when retrieval
helps vs.\ when parametric memory
suffices~\citep{mallen2023whennot}, quantified the relative
contribution of context vs.\ parameters in long-form
generation~\citep{tao2024whencontext}, and used entity
substitution to force context
reading~\citep{longpre2021entitybased}.
MIRAGE~\citep{xiong2024mirage} reports a no-retrieval
baseline alongside RAG accuracy on medical QA.

\section{Appendix - Boxoffice}
\label{app:bo-generation}
\subsection{Example Prompt Components}
\label{app:bo-example-prompt}

\noindent
We show below the exact prompt components used by the  \bo{} seeds: the instruction/system prompt chunk, one example of movie
dossier chunk, and the final question suffix. The text is taken verbatim from a \texttt{s7} seed row.
Each query includes the instruction chunk, then 10 dossier chunk, and finally the question suffix.

\begin{center}
\fcolorbox{black!15}{black!5}{%
\parbox{0.94\linewidth}{%
\footnotesize
\textbf{Instruction chunk} \\
\ttfamily
Use only the synthetic movie dossiers below. Ignore outside/world knowledge. \\
For ranking, compare only the BOX\_OFFICE\_MUSD integer fields across all named \\
candidates. Return exactly one FILM-ID and no other text.}}
\end{center}

\begin{center}
\fcolorbox{black!15}{black!5}{%
\parbox{0.94\linewidth}{%
\scriptsize
\textbf{Example dossier chunk} \\
\ttfamily
BENCHMARK\_DOSSIER \\
ENTITY\_ID: FILM-1008 \\
TITLE: The Midnight Ledger \\
DIRECTOR: Elara Vance \\
RELEASE\_YEAR: 2009 \\
STARLIGHT\_AWARDS: 4 \\
BOX\_OFFICE\_MUSD: 966 \\
GENRE: Sci-Fi \\
STUDIO: Aperture Seven \\
COUNTRY: Norland \\
RUNTIME\_MIN: 142 \\
CAST: Lyra Boone, Sage Monroe, Rin Carter \\
\\
CATALOG\_ENTRY \\
Overview: The Midnight Ledger is a 2009 sci-fi feature directed by Elara \\
Vance for Aperture Seven. In short directory listings, the film is usually \\
identified through its Norland market line and its association with Novastad. \\
Cast and setting: The credited cast is Lyra Boone, Sage Monroe, Rin Carter. \\
Reference copy keeps Novastad attached to the title as its standing location \\
marker rather than expanding into a long plot synopsis. \\
Release profile: The entry presents the film as a full-length commercial \\
feature with a runtime of 142 minutes. The dossier keeps the release identity \\
tied to studio, market, cast, and setting in the style of a compact film \\
guide. \\
Commercial note: The recorded box office for The Midnight Ledger is 966 million \\
USD, and the listing treats that figure as part of the title's market profile \\
rather than as a review note. \\
\\
RELEASE\_AND\_REFERENCE\_NOTES \\
SETTING\_CITY: Novastad \\
SETTING\_CITY\_POPULATION\_MIL: 9.2 \\
Studio filing: Aperture Seven remains the credited company throughout the \\
record. \\
Market filing: Norland is the territory used for release and indexing \\
reference. \\
Cast filing: Lyra Boone, Sage Monroe, Rin Carter remains the recurring cast \\
line attached to the title. \\
Setting filing: Novastad is the place-name consistently used to anchor the \\
film in short reference prose. \\
Directory note: The title is grouped with current genre features rather than \\
with repertory or specialty listings. \\
Exhibition note: The entry reads like a film associated with standard \\
multiplex play. \\
Reference note: The record keeps to production identity, market placement, \\
cast, setting, runtime, and gross without adding review language. \\
Index note: The Midnight Ledger is filed in a director-led pattern that makes \\
the title easy to locate in long title lists. \\
Program note: The note block favors release and reference detail over plot \\
description. \\
Library note: The overall effect is closer to a compact distribution-reference \\
record than to a pressbook or capsule review. \\
Release note: The title is presented as a studio-backed sci-fi feature rather \\
than as a limited event item. \\
Record line: The runtime and release year remain part of the short production \\
summary for the film.}}
\end{center}

\begin{center}
\fcolorbox{black!15}{black!5}{%
\parbox{0.94\linewidth}{%
\footnotesize
\textbf{Question suffix} \\
\ttfamily
Question: Valid candidates: FILM-1008, FILM-1022, FILM-1002, FILM-2000, \\
FILM-2001, FILM-1000, FILM-1044, FILM-2002, FILM-1032, FILM-2003 \\
Which valid FILM-ID has the maximum BOX\_OFFICE\_MUSD among all listed \\
candidates? \\
Read the BOX\_OFFICE\_MUSD field from the dossiers; do not infer values from \\
titles. \\
Return exactly one valid FILM-ID.}}
\end{center}

\subsection{Warmup configurations}
\begin{table}[h]
\scriptsize
\renewcommand{\arraystretch}{1.1}
\setlength{\tabcolsep}{4pt}
\caption{Configurations of a \bo{} query, by the role each chunk
played during warmup ($W$inner vs $N$onwinner).}
\label{tab:cells}
\hspace{-18pt}
\begin{tabular}{l c c p{0.65\linewidth}}
\toprule
Configuration & Winner was & Nonwinners were& Description \\
\midrule
\texttt{aligned} & $W$ & $N$ &
  Every chunk's cached state matches its role in the eval query.
  The winner's cache encodes ``I am the maximum''; each nonwinner's
  cache encodes ``I am not''. \textbf{Easy case.} \\
\addlinespace[2pt]
\texttt{all-W}   & $W$ & $W$ &
  The winner's cache is aligned, but each nonwinner's cache also
  encodes ``I am the maximum''. The model must override the 
  ``winner'' signal carried by nonwinner caches.
  \textbf{Medium case.} \\
\addlinespace[2pt]
\texttt{all-N}   & $N$ & $N$ &
  Every nonwinner's cache is aligned, but the winner's 
  encodes ``I am not the maximum''. The model must identify the
  winner despite every cache claiming nonwinner status.
  \textbf{Medium case.} \\
\addlinespace[2pt]
\texttt{flipped} & $N$ & $W$ &
  The signals are fully inverted relative to the eval roles. The
  model must recover the correct ranking from a reversed set of
  cached signals. \textbf{Hard case.} \\
\bottomrule
\end{tabular}
\end{table}
Table~\ref{tab:cells} lists the four warmup configurations of Section~\ref{sec:boxoffice}, the cached signal each presents to the reuse method, and the qualitative difficulty it imposes.

\subsection{Dataset generation}
\noindent\textbf{Pseudocode for generating one \bo{} seed.}
Algorithm 1 summarizes the workflow implemented in the
released \bo{} bundle for a single seed $s$.
~\\

{\small
\noindent\textbf{Algorithm 1: Generate one \bo{} seed $s$}
\medskip
\noindent\textbf{Input:} seed $s$; film corpus; similarity metadata $W$;
filter model set $\mathcal{M}$

\noindent\textbf{Output:} \texttt{full.jsonl}, \texttt{eval.jsonl},
manifest, validation counts

\medskip
\begin{algorithmic}[1]
\State rebuild similarity metadata $W$ from the corpus if missing
\State load film records; build warmup schedule with W/N roles
\ForAll{$c \in \{\texttt{aligned},\texttt{all-W},\texttt{all-N},\texttt{flipped}\}$}
  \State oversample candidate 10-film queries using seed $s$
  \State compute similarity statistics on candidates via $W$
  \State retain retrieval-balanced candidate pool for $c$
\EndFor
\State extract eval rows; build question-only copies
\ForAll{$m \in \mathcal{M}$}
  \State run full-context and question-only baselines on eval rows
\EndFor
\State keep rows with full-context $F_1{=}1$ and question-only
       $F_1{\neq}1$ for \emph{every} $m\in\mathcal{M}$
\State per configuration, retain a fixed number of surviving rows
\State augment warmup so every eval chunk has $\geq 2$ cached
       \emph{smaller-than} and $\geq 2$ \emph{greater-than}
       predecessor configurations
\State concatenate augmented warmup with filtered eval; reindex
\State write \texttt{full.jsonl}, \texttt{eval.jsonl}, manifest,
       and validation counts
\end{algorithmic}
}

Steps~9--12 implement the  selection that preserves the ``meaningful by construction'' property by keeping
only rows that all filter models solve with full context but not without
context. Step~14 realizes the directional warmup balancing discussed in
Section~\ref{sec:boxoffice}: every evaluation chunk is given both
``smaller-than'' and ``greater-than'' cached predecessors, which yields the
balanced seeds used in the paper.

\subsection{Chunk distance CDFs}\label{appendix:cdf}
\begin{figure}[t]
\centering
\includegraphics[width=\linewidth]{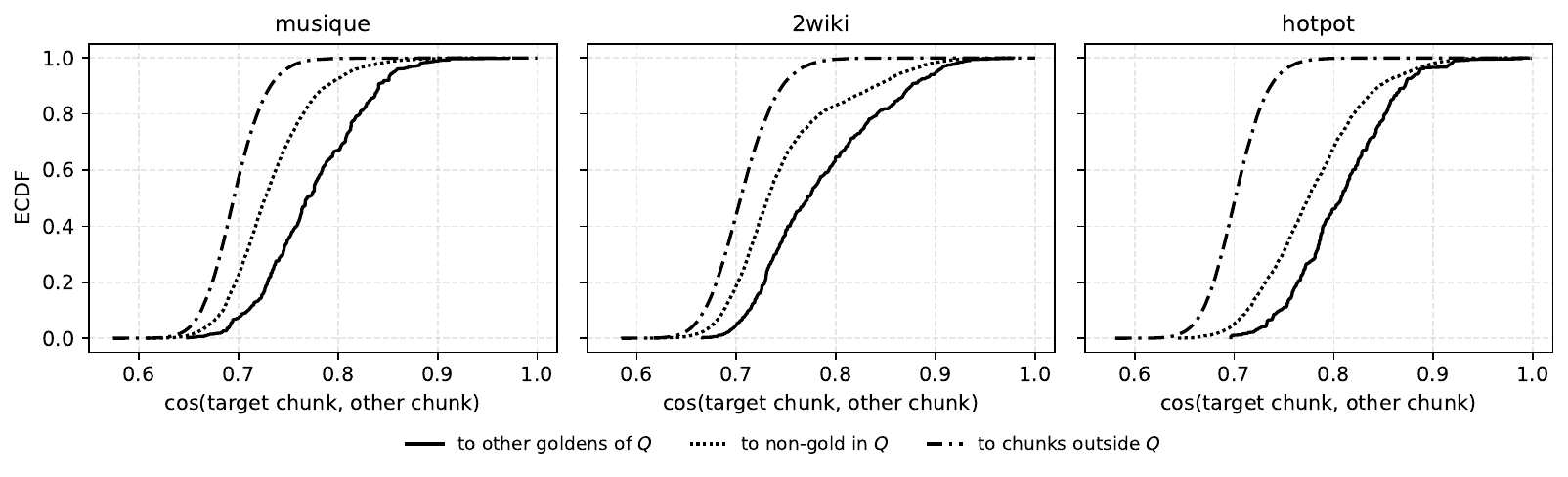}\\[0.3em]
\includegraphics[width=\linewidth]{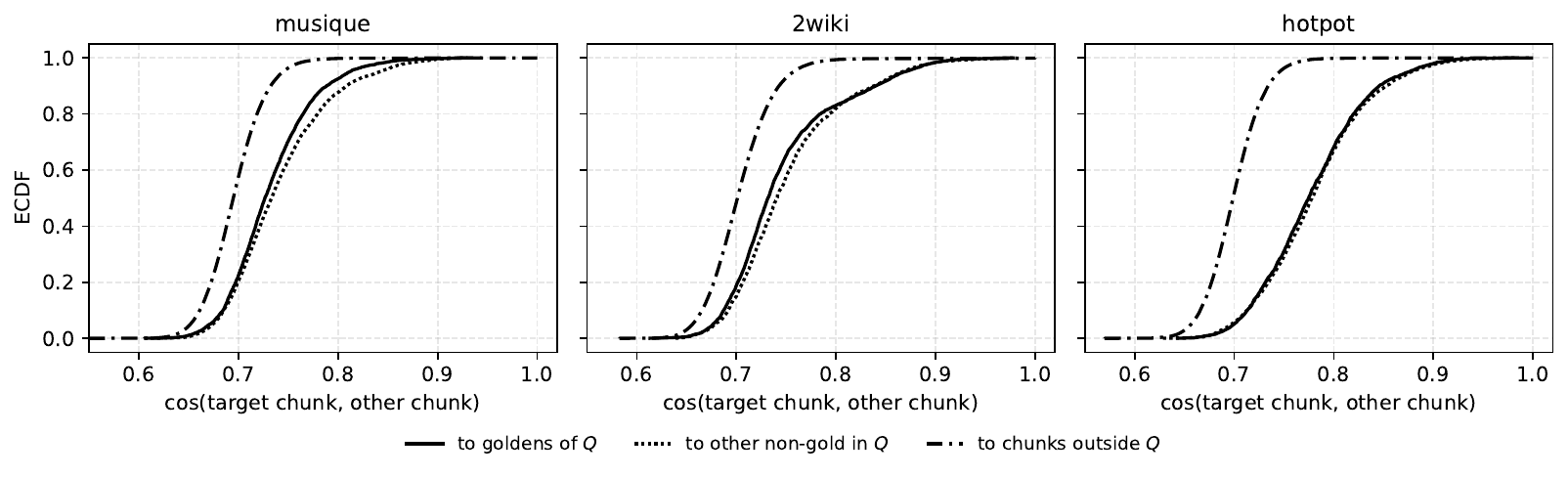}\\[0.3em]
\includegraphics[width=0.65\linewidth]{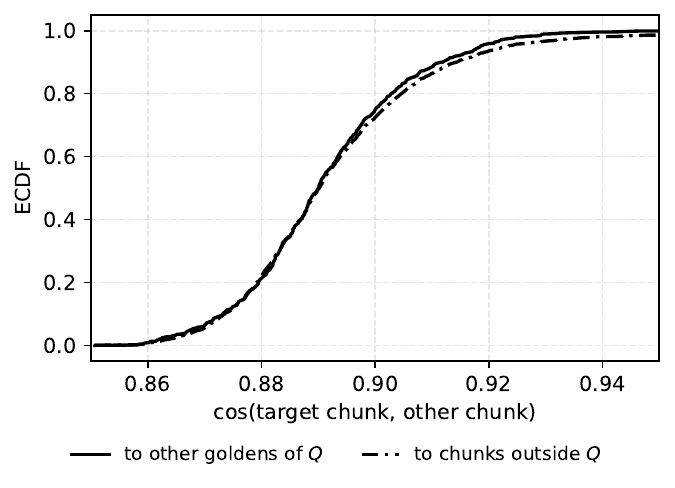}
\caption{{\bf Top}: similarity between a {\em golden} target chunk in query Q of LongBench-QA w.r.t. other golden chunks Q, other non-golden chunks in Q, and other chunks in the dataset. {\bf Middle}: same as top but done for target {\em non-golden} chunks. {\bf Bottom}: similarity between a chunk in \bo{} in Q w.r.t. other chunks in Q  and other chunks in the dataset.}
\label{fig:dist-cdf}
\end{figure}

Figure~\ref{fig:dist-cdf} shows the distribution of the distances between chunk in the context of a query Q and other chunks in the context of Q or outside of Q. The main result is that in LongBench-QA golden chunks are clearly closer to other chunks in the same query, and especially golden (top). A non-golden chunk that appear in Q is also more close to other chunks Q than to other chunks. Hence, LongBench-QA is highly clustered. On the contrary, chunks in \bo{} (which are all golden ones) have the same similarity distribution towards any other chunk in the dataset.

\subsection{\bo{} additional evaluation}
\label{appendix:bo}

Table~\ref{tab:app:bo-15} extends the results of Section~\ref{sec:boxoffice:results} for $R=0.15$ with standard deviation information per cell, computed on the results of 10 seeds. Table~\ref{tab:app:bo-05} the result obtained on \bo{} for $R=0.05$.
The most interesting result is that at $R=0.05$, the winner on average for each model is $LM+Q$. The reason is that at $R=0.05$ the information encoded in the KV cache during warmup cannot be adequately overwritten by means of selective recomputation, and hence decisively influence the accuracy achieved. $LM+Q$ achieves near-perfect accuracy on the \texttt{aligned} cells, which skews the average results.  

We report that  running an end-to-end experiment from dataset generation to evaluation on all the involved methods takes roughly 5 hours on an accelerator with
$\sim$376~TFLOPS BF16 dense compute, 64\,GB HBM2E, and
$\sim$1.6\,TB/s memory bandwidth per device.
\begin{table}[t]
\centering
\scriptsize
\renewcommand{\arraystretch}{0.95}
\setlength{\tabcolsep}{2pt}
\caption{Norm-F1 on \bo{} at $R{=}0.15$, mean${\pm}$sample std across 10 generated datasets (seeds $7,11,13,17,19,23,29,31,47,73$). Within each (model, configuration), the winning method is in bold.}
\label{tab:app:bo-15}
\begin{tabular}{c|l *{10}{c}}
\toprule
 & Config
& \rotatebox{90}{\,CB} & \rotatebox{90}{\,CB+Q}
& \rotatebox{90}{\,FR} & \rotatebox{90}{\,FR+Q}
& \rotatebox{90}{\,LM} & \rotatebox{90}{\,LM+Q}
& \rotatebox{90}{\,CC-M2} & \rotatebox{90}{\,CC-M2+Q}
& \rotatebox{90}{\,CC-M4} & \rotatebox{90}{\,CC-M4+Q} \\
\midrule
\multirow{5}{*}{\rotatebox[origin=c]{90}{\textbf{Qwen}}}
 & aligned & $.39{\scriptscriptstyle\,\pm\,.07}$ & $.09{\scriptscriptstyle\,\pm\,.05}$ & $.44{\scriptscriptstyle\,\pm\,.06}$ & $.40{\scriptscriptstyle\,\pm\,.10}$ & $\mathbf{.97{\scriptscriptstyle\,\pm\,.03}}$ & $.94{\scriptscriptstyle\,\pm\,.05}$ & $.17{\scriptscriptstyle\,\pm\,.13}$ & $.38{\scriptscriptstyle\,\pm\,.11}$ & $.18{\scriptscriptstyle\,\pm\,.09}$ & $.25{\scriptscriptstyle\,\pm\,.10}$ \\
 & all-W & $.30{\scriptscriptstyle\,\pm\,.06}$ & $.12{\scriptscriptstyle\,\pm\,.06}$ & $.28{\scriptscriptstyle\,\pm\,.07}$ & $.29{\scriptscriptstyle\,\pm\,.09}$ & $.11{\scriptscriptstyle\,\pm\,.06}$ & $.25{\scriptscriptstyle\,\pm\,.05}$ & $.13{\scriptscriptstyle\,\pm\,.10}$ & $.27{\scriptscriptstyle\,\pm\,.08}$ & $.24{\scriptscriptstyle\,\pm\,.11}$ & $\mathbf{.32{\scriptscriptstyle\,\pm\,.07}}$ \\
 & all-N & $.39{\scriptscriptstyle\,\pm\,.06}$ & $.20{\scriptscriptstyle\,\pm\,.07}$ & $.30{\scriptscriptstyle\,\pm\,.06}$ & $\mathbf{.39{\scriptscriptstyle\,\pm\,.11}}$ & $.05{\scriptscriptstyle\,\pm\,.03}$ & $.11{\scriptscriptstyle\,\pm\,.03}$ & $.18{\scriptscriptstyle\,\pm\,.09}$ & $.36{\scriptscriptstyle\,\pm\,.11}$ & $.24{\scriptscriptstyle\,\pm\,.09}$ & $.34{\scriptscriptstyle\,\pm\,.07}$ \\
 & flipped & $\mathbf{.41{\scriptscriptstyle\,\pm\,.12}}$ & $.18{\scriptscriptstyle\,\pm\,.07}$ & $.23{\scriptscriptstyle\,\pm\,.07}$ & $.28{\scriptscriptstyle\,\pm\,.06}$ & $.00{\scriptscriptstyle\,\pm\,.00}$ & $.06{\scriptscriptstyle\,\pm\,.04}$ & $.15{\scriptscriptstyle\,\pm\,.08}$ & $.29{\scriptscriptstyle\,\pm\,.09}$ & $.28{\scriptscriptstyle\,\pm\,.11}$ & $.34{\scriptscriptstyle\,\pm\,.10}$ \\
\cmidrule(lr){2-12}
 & average & $\mathbf{.37{\scriptscriptstyle\,\pm\,.05}}$ & $.15{\scriptscriptstyle\,\pm\,.03}$ & $.31{\scriptscriptstyle\,\pm\,.04}$ & $.34{\scriptscriptstyle\,\pm\,.04}$ & $.28{\scriptscriptstyle\,\pm\,.02}$ & $.34{\scriptscriptstyle\,\pm\,.02}$ & $.16{\scriptscriptstyle\,\pm\,.05}$ & $.33{\scriptscriptstyle\,\pm\,.05}$ & $.24{\scriptscriptstyle\,\pm\,.04}$ & $.31{\scriptscriptstyle\,\pm\,.04}$ \\
\midrule
\multirow{5}{*}{\rotatebox[origin=c]{90}{\textbf{Llama}}}
 & aligned & $.10{\scriptscriptstyle\,\pm\,.03}$ & $.31{\scriptscriptstyle\,\pm\,.09}$ & $.11{\scriptscriptstyle\,\pm\,.05}$ & $.37{\scriptscriptstyle\,\pm\,.07}$ & $\mathbf{.98{\scriptscriptstyle\,\pm\,.02}}$ & $.95{\scriptscriptstyle\,\pm\,.03}$ & $.17{\scriptscriptstyle\,\pm\,.11}$ & $.55{\scriptscriptstyle\,\pm\,.11}$ & $.19{\scriptscriptstyle\,\pm\,.06}$ & $.51{\scriptscriptstyle\,\pm\,.08}$ \\
 & all-W & $.06{\scriptscriptstyle\,\pm\,.04}$ & $.21{\scriptscriptstyle\,\pm\,.03}$ & $.05{\scriptscriptstyle\,\pm\,.04}$ & $.28{\scriptscriptstyle\,\pm\,.06}$ & $.14{\scriptscriptstyle\,\pm\,.08}$ & $.29{\scriptscriptstyle\,\pm\,.07}$ & $.13{\scriptscriptstyle\,\pm\,.11}$ & $.42{\scriptscriptstyle\,\pm\,.12}$ & $.24{\scriptscriptstyle\,\pm\,.09}$ & $\mathbf{.50{\scriptscriptstyle\,\pm\,.12}}$ \\
 & all-N & $.06{\scriptscriptstyle\,\pm\,.05}$ & $.25{\scriptscriptstyle\,\pm\,.12}$ & $.13{\scriptscriptstyle\,\pm\,.06}$ & $.43{\scriptscriptstyle\,\pm\,.08}$ & $.07{\scriptscriptstyle\,\pm\,.04}$ & $.55{\scriptscriptstyle\,\pm\,.06}$ & $.16{\scriptscriptstyle\,\pm\,.10}$ & $.57{\scriptscriptstyle\,\pm\,.13}$ & $.23{\scriptscriptstyle\,\pm\,.09}$ & $\mathbf{.57{\scriptscriptstyle\,\pm\,.10}}$ \\
 & flipped & $.09{\scriptscriptstyle\,\pm\,.04}$ & $.25{\scriptscriptstyle\,\pm\,.08}$ & $.06{\scriptscriptstyle\,\pm\,.04}$ & $.35{\scriptscriptstyle\,\pm\,.08}$ & $.00{\scriptscriptstyle\,\pm\,.00}$ & $.09{\scriptscriptstyle\,\pm\,.06}$ & $.11{\scriptscriptstyle\,\pm\,.08}$ & $.41{\scriptscriptstyle\,\pm\,.13}$ & $.23{\scriptscriptstyle\,\pm\,.11}$ & $\mathbf{.54{\scriptscriptstyle\,\pm\,.10}}$ \\
\cmidrule(lr){2-12}
 & average & $.08{\scriptscriptstyle\,\pm\,.03}$ & $.26{\scriptscriptstyle\,\pm\,.04}$ & $.09{\scriptscriptstyle\,\pm\,.02}$ & $.36{\scriptscriptstyle\,\pm\,.03}$ & $.30{\scriptscriptstyle\,\pm\,.03}$ & $.47{\scriptscriptstyle\,\pm\,.03}$ & $.14{\scriptscriptstyle\,\pm\,.05}$ & $.49{\scriptscriptstyle\,\pm\,.07}$ & $.22{\scriptscriptstyle\,\pm\,.04}$ & $\mathbf{.53{\scriptscriptstyle\,\pm\,.06}}$ \\
\midrule
\multirow{5}{*}{\rotatebox[origin=c]{90}{\textbf{Mistral}}}
 & aligned & $.20{\scriptscriptstyle\,\pm\,.08}$ & $.18{\scriptscriptstyle\,\pm\,.08}$ & $.33{\scriptscriptstyle\,\pm\,.06}$ & $.53{\scriptscriptstyle\,\pm\,.10}$ & $\mathbf{.98{\scriptscriptstyle\,\pm\,.02}}$ & $.94{\scriptscriptstyle\,\pm\,.03}$ & $.14{\scriptscriptstyle\,\pm\,.10}$ & $.42{\scriptscriptstyle\,\pm\,.13}$ & $.21{\scriptscriptstyle\,\pm\,.08}$ & $.36{\scriptscriptstyle\,\pm\,.10}$ \\
 & all-W & $.24{\scriptscriptstyle\,\pm\,.06}$ & $.15{\scriptscriptstyle\,\pm\,.07}$ & $.15{\scriptscriptstyle\,\pm\,.08}$ & $.26{\scriptscriptstyle\,\pm\,.06}$ & $.17{\scriptscriptstyle\,\pm\,.05}$ & $.26{\scriptscriptstyle\,\pm\,.09}$ & $.14{\scriptscriptstyle\,\pm\,.09}$ & $.26{\scriptscriptstyle\,\pm\,.08}$ & $.31{\scriptscriptstyle\,\pm\,.09}$ & $\mathbf{.34{\scriptscriptstyle\,\pm\,.07}}$ \\
 & all-N & $.29{\scriptscriptstyle\,\pm\,.08}$ & $.16{\scriptscriptstyle\,\pm\,.07}$ & $.24{\scriptscriptstyle\,\pm\,.07}$ & $\mathbf{.57{\scriptscriptstyle\,\pm\,.09}}$ & $.09{\scriptscriptstyle\,\pm\,.04}$ & $.29{\scriptscriptstyle\,\pm\,.09}$ & $.16{\scriptscriptstyle\,\pm\,.08}$ & $.34{\scriptscriptstyle\,\pm\,.15}$ & $.32{\scriptscriptstyle\,\pm\,.15}$ & $.39{\scriptscriptstyle\,\pm\,.14}$ \\
 & flipped & $.25{\scriptscriptstyle\,\pm\,.09}$ & $.11{\scriptscriptstyle\,\pm\,.05}$ & $.18{\scriptscriptstyle\,\pm\,.05}$ & $\mathbf{.34{\scriptscriptstyle\,\pm\,.10}}$ & $.00{\scriptscriptstyle\,\pm\,.00}$ & $.02{\scriptscriptstyle\,\pm\,.02}$ & $.13{\scriptscriptstyle\,\pm\,.09}$ & $.23{\scriptscriptstyle\,\pm\,.14}$ & $.30{\scriptscriptstyle\,\pm\,.13}$ & $.30{\scriptscriptstyle\,\pm\,.10}$ \\
\cmidrule(lr){2-12}
 & average & $.24{\scriptscriptstyle\,\pm\,.03}$ & $.15{\scriptscriptstyle\,\pm\,.04}$ & $.23{\scriptscriptstyle\,\pm\,.03}$ & $\mathbf{.42{\scriptscriptstyle\,\pm\,.04}}$ & $.31{\scriptscriptstyle\,\pm\,.02}$ & $.38{\scriptscriptstyle\,\pm\,.04}$ & $.14{\scriptscriptstyle\,\pm\,.05}$ & $.31{\scriptscriptstyle\,\pm\,.06}$ & $.28{\scriptscriptstyle\,\pm\,.07}$ & $.35{\scriptscriptstyle\,\pm\,.05}$ \\
\bottomrule
\end{tabular}
\end{table}

\begin{table}[t]
\centering
\scriptsize
\renewcommand{\arraystretch}{0.95}
\setlength{\tabcolsep}{2pt}
\caption{norm-F1 on \bo{} at $R{=}0.05$, mean${\pm}$sample std across 10 generated datasets (seeds $7,11,13,17,19,23,29,31,47,73$). Within each (model, configuration), the winning method is in bold.}
\label{tab:app:bo-05}
\begin{tabular}{c|l *{10}{c}}
\toprule
 & Config
& \rotatebox{90}{\,CB} & \rotatebox{90}{\,CB+Q}
& \rotatebox{90}{\,FR} & \rotatebox{90}{\,FR+Q}
& \rotatebox{90}{\,LM} & \rotatebox{90}{\,LM+Q}
& \rotatebox{90}{\,CC-M2} & \rotatebox{90}{\,CC-M2+Q}
& \rotatebox{90}{\,CC-M4} & \rotatebox{90}{\,CC-M4+Q} \\
\midrule
\multirow{5}{*}{\rotatebox[origin=c]{90}{\textbf{Qwen}}}
 & aligned & $.03{\scriptscriptstyle\,\pm\,.02}$ & $.07{\scriptscriptstyle\,\pm\,.04}$ & $.36{\scriptscriptstyle\,\pm\,.05}$ & $.34{\scriptscriptstyle\,\pm\,.06}$ & $.95{\scriptscriptstyle\,\pm\,.02}$ & $\mathbf{.97{\scriptscriptstyle\,\pm\,.02}}$ & $.13{\scriptscriptstyle\,\pm\,.13}$ & $.18{\scriptscriptstyle\,\pm\,.12}$ & $.12{\scriptscriptstyle\,\pm\,.07}$ & $.10{\scriptscriptstyle\,\pm\,.05}$ \\
 & all-W & $.04{\scriptscriptstyle\,\pm\,.03}$ & $.07{\scriptscriptstyle\,\pm\,.03}$ & $\mathbf{.23{\scriptscriptstyle\,\pm\,.05}}$ & $.23{\scriptscriptstyle\,\pm\,.04}$ & $.11{\scriptscriptstyle\,\pm\,.05}$ & $.18{\scriptscriptstyle\,\pm\,.05}$ & $.10{\scriptscriptstyle\,\pm\,.07}$ & $.12{\scriptscriptstyle\,\pm\,.08}$ & $.21{\scriptscriptstyle\,\pm\,.08}$ & $.20{\scriptscriptstyle\,\pm\,.06}$ \\
 & all-N & $.07{\scriptscriptstyle\,\pm\,.04}$ & $.09{\scriptscriptstyle\,\pm\,.04}$ & $\mathbf{.28{\scriptscriptstyle\,\pm\,.04}}$ & $.28{\scriptscriptstyle\,\pm\,.06}$ & $.04{\scriptscriptstyle\,\pm\,.03}$ & $.08{\scriptscriptstyle\,\pm\,.03}$ & $.14{\scriptscriptstyle\,\pm\,.08}$ & $.15{\scriptscriptstyle\,\pm\,.07}$ & $.15{\scriptscriptstyle\,\pm\,.07}$ & $.18{\scriptscriptstyle\,\pm\,.09}$ \\
 & flipped & $.11{\scriptscriptstyle\,\pm\,.06}$ & $.11{\scriptscriptstyle\,\pm\,.06}$ & $.19{\scriptscriptstyle\,\pm\,.05}$ & $\mathbf{.19{\scriptscriptstyle\,\pm\,.05}}$ & $.00{\scriptscriptstyle\,\pm\,.00}$ & $.00{\scriptscriptstyle\,\pm\,.00}$ & $.12{\scriptscriptstyle\,\pm\,.08}$ & $.14{\scriptscriptstyle\,\pm\,.07}$ & $.19{\scriptscriptstyle\,\pm\,.09}$ & $.19{\scriptscriptstyle\,\pm\,.08}$ \\
\cmidrule(lr){2-12}
 & average & $.07{\scriptscriptstyle\,\pm\,.02}$ & $.09{\scriptscriptstyle\,\pm\,.02}$ & $.27{\scriptscriptstyle\,\pm\,.03}$ & $.26{\scriptscriptstyle\,\pm\,.02}$ & $.28{\scriptscriptstyle\,\pm\,.02}$ & $\mathbf{.31{\scriptscriptstyle\,\pm\,.01}}$ & $.12{\scriptscriptstyle\,\pm\,.04}$ & $.15{\scriptscriptstyle\,\pm\,.04}$ & $.17{\scriptscriptstyle\,\pm\,.03}$ & $.17{\scriptscriptstyle\,\pm\,.03}$ \\
\midrule
\multirow{5}{*}{\rotatebox[origin=c]{90}{\textbf{Llama}}}
 & aligned & $.11{\scriptscriptstyle\,\pm\,.06}$ & $.12{\scriptscriptstyle\,\pm\,.06}$ & $.09{\scriptscriptstyle\,\pm\,.03}$ & $.09{\scriptscriptstyle\,\pm\,.05}$ & $\mathbf{.98{\scriptscriptstyle\,\pm\,.02}}$ & $.96{\scriptscriptstyle\,\pm\,.03}$ & $.13{\scriptscriptstyle\,\pm\,.11}$ & $.21{\scriptscriptstyle\,\pm\,.11}$ & $.10{\scriptscriptstyle\,\pm\,.04}$ & $.16{\scriptscriptstyle\,\pm\,.07}$ \\
 & all-W & $.04{\scriptscriptstyle\,\pm\,.04}$ & $.06{\scriptscriptstyle\,\pm\,.03}$ & $.04{\scriptscriptstyle\,\pm\,.03}$ & $.05{\scriptscriptstyle\,\pm\,.03}$ & $.13{\scriptscriptstyle\,\pm\,.08}$ & $.18{\scriptscriptstyle\,\pm\,.05}$ & $.11{\scriptscriptstyle\,\pm\,.10}$ & $.15{\scriptscriptstyle\,\pm\,.10}$ & $.19{\scriptscriptstyle\,\pm\,.09}$ & $\mathbf{.22{\scriptscriptstyle\,\pm\,.11}}$ \\
 & all-N & $.11{\scriptscriptstyle\,\pm\,.05}$ & $.08{\scriptscriptstyle\,\pm\,.04}$ & $.15{\scriptscriptstyle\,\pm\,.06}$ & $.12{\scriptscriptstyle\,\pm\,.05}$ & $.04{\scriptscriptstyle\,\pm\,.03}$ & $\mathbf{.46{\scriptscriptstyle\,\pm\,.08}}$ & $.14{\scriptscriptstyle\,\pm\,.10}$ & $.21{\scriptscriptstyle\,\pm\,.11}$ & $.15{\scriptscriptstyle\,\pm\,.09}$ & $.20{\scriptscriptstyle\,\pm\,.09}$ \\
 & flipped & $.04{\scriptscriptstyle\,\pm\,.03}$ & $.07{\scriptscriptstyle\,\pm\,.04}$ & $.05{\scriptscriptstyle\,\pm\,.03}$ & $.05{\scriptscriptstyle\,\pm\,.03}$ & $.00{\scriptscriptstyle\,\pm\,.00}$ & $.01{\scriptscriptstyle\,\pm\,.01}$ & $.07{\scriptscriptstyle\,\pm\,.04}$ & $.14{\scriptscriptstyle\,\pm\,.08}$ & $.16{\scriptscriptstyle\,\pm\,.08}$ & $\mathbf{.19{\scriptscriptstyle\,\pm\,.09}}$ \\
\cmidrule(lr){2-12}
 & average & $.07{\scriptscriptstyle\,\pm\,.03}$ & $.08{\scriptscriptstyle\,\pm\,.02}$ & $.08{\scriptscriptstyle\,\pm\,.02}$ & $.08{\scriptscriptstyle\,\pm\,.02}$ & $.29{\scriptscriptstyle\,\pm\,.03}$ & $\mathbf{.40{\scriptscriptstyle\,\pm\,.03}}$ & $.11{\scriptscriptstyle\,\pm\,.04}$ & $.18{\scriptscriptstyle\,\pm\,.04}$ & $.15{\scriptscriptstyle\,\pm\,.03}$ & $.19{\scriptscriptstyle\,\pm\,.04}$ \\
\midrule
\multirow{5}{*}{\rotatebox[origin=c]{90}{\textbf{Mistral}}}
 & aligned & $.07{\scriptscriptstyle\,\pm\,.05}$ & $.09{\scriptscriptstyle\,\pm\,.04}$ & $.35{\scriptscriptstyle\,\pm\,.07}$ & $.34{\scriptscriptstyle\,\pm\,.09}$ & $.98{\scriptscriptstyle\,\pm\,.02}$ & $\mathbf{.98{\scriptscriptstyle\,\pm\,.02}}$ & $.11{\scriptscriptstyle\,\pm\,.10}$ & $.23{\scriptscriptstyle\,\pm\,.11}$ & $.17{\scriptscriptstyle\,\pm\,.06}$ & $.19{\scriptscriptstyle\,\pm\,.08}$ \\
 & all-W & $.05{\scriptscriptstyle\,\pm\,.04}$ & $.05{\scriptscriptstyle\,\pm\,.03}$ & $.13{\scriptscriptstyle\,\pm\,.06}$ & $.15{\scriptscriptstyle\,\pm\,.06}$ & $.17{\scriptscriptstyle\,\pm\,.06}$ & $.22{\scriptscriptstyle\,\pm\,.08}$ & $.11{\scriptscriptstyle\,\pm\,.08}$ & $.14{\scriptscriptstyle\,\pm\,.08}$ & $.23{\scriptscriptstyle\,\pm\,.07}$ & $\mathbf{.25{\scriptscriptstyle\,\pm\,.09}}$ \\
 & all-N & $.08{\scriptscriptstyle\,\pm\,.02}$ & $.17{\scriptscriptstyle\,\pm\,.05}$ & $.21{\scriptscriptstyle\,\pm\,.04}$ & $\mathbf{.26{\scriptscriptstyle\,\pm\,.06}}$ & $.04{\scriptscriptstyle\,\pm\,.02}$ & $.14{\scriptscriptstyle\,\pm\,.04}$ & $.13{\scriptscriptstyle\,\pm\,.10}$ & $.20{\scriptscriptstyle\,\pm\,.12}$ & $.23{\scriptscriptstyle\,\pm\,.11}$ & $.24{\scriptscriptstyle\,\pm\,.14}$ \\
 & flipped & $.03{\scriptscriptstyle\,\pm\,.01}$ & $.04{\scriptscriptstyle\,\pm\,.03}$ & $.16{\scriptscriptstyle\,\pm\,.04}$ & $.19{\scriptscriptstyle\,\pm\,.05}$ & $.00{\scriptscriptstyle\,\pm\,.00}$ & $.00{\scriptscriptstyle\,\pm\,.01}$ & $.10{\scriptscriptstyle\,\pm\,.06}$ & $.14{\scriptscriptstyle\,\pm\,.09}$ & $\mathbf{.22{\scriptscriptstyle\,\pm\,.10}}$ & $.21{\scriptscriptstyle\,\pm\,.10}$ \\
\cmidrule(lr){2-12}
 & average & $.06{\scriptscriptstyle\,\pm\,.02}$ & $.09{\scriptscriptstyle\,\pm\,.02}$ & $.21{\scriptscriptstyle\,\pm\,.02}$ & $.23{\scriptscriptstyle\,\pm\,.03}$ & $.29{\scriptscriptstyle\,\pm\,.02}$ & $\mathbf{.33{\scriptscriptstyle\,\pm\,.03}}$ & $.11{\scriptscriptstyle\,\pm\,.04}$ & $.18{\scriptscriptstyle\,\pm\,.05}$ & $.21{\scriptscriptstyle\,\pm\,.05}$ & $.22{\scriptscriptstyle\,\pm\,.06}$ \\
\bottomrule
\end{tabular}
\end{table}

\begin{figure}[htbp]
\centering
\includegraphics[width=\linewidth]{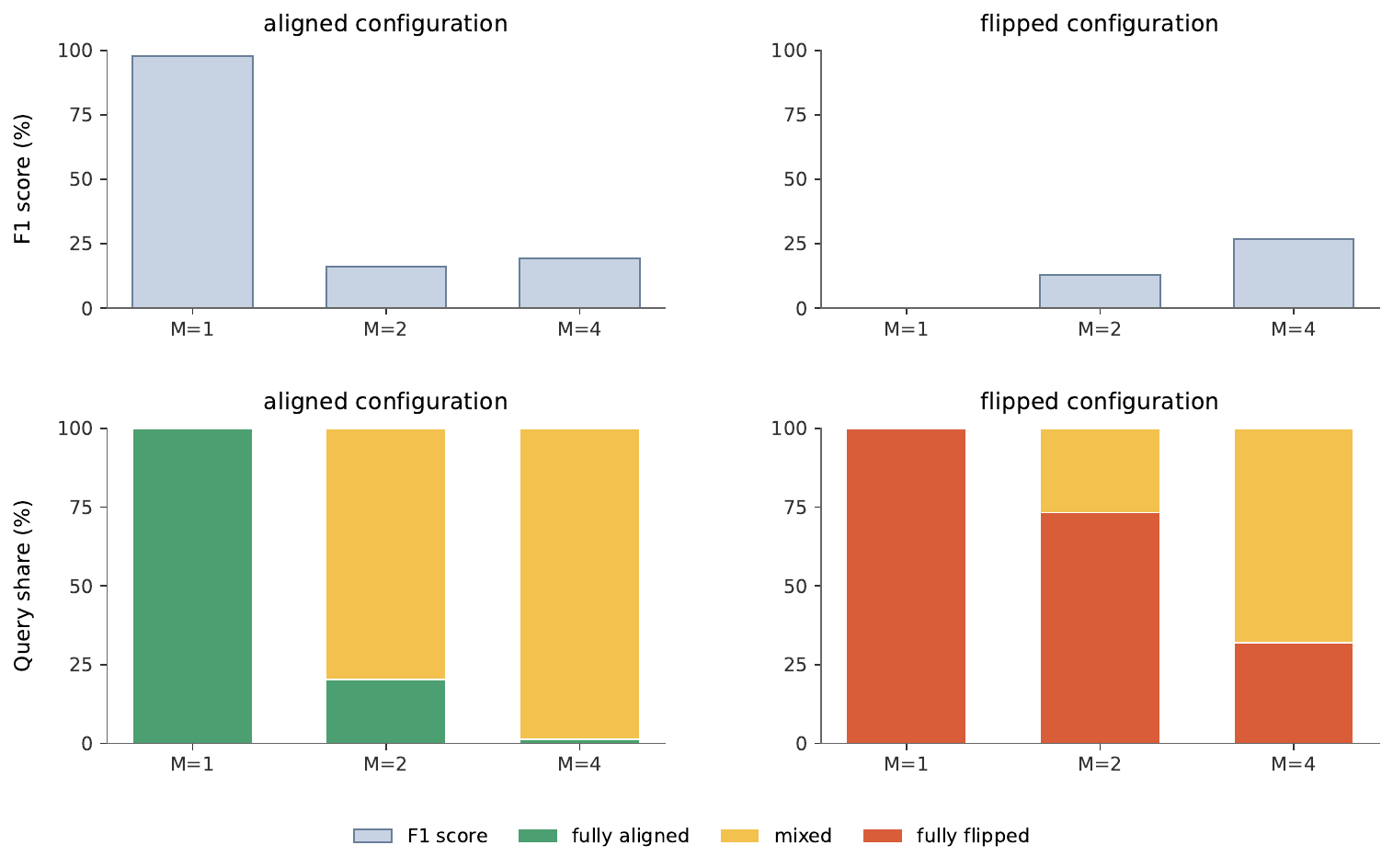}
\caption{LMCache ($M=1$) and CacheCraft ($M=2,4)$ performance and query-level orientation breakdown on the ten \bo{} seeds at recompute ratio $R{=}0.15$. The top row shows mean F1 across all three models for the aligned  and flipped  configurations as the number of
  stored variants grows from $M{=}1$ to $M{=}4$. The bottom row classifies each evaluation query by the orientation of the selected versions across all 10 chunks: {\em fully aligned} means every selected winner/nonwinner version has the role-consistent orientation, {\em
  mixed} means the query contains both properly oriented (aligned) and misoriented (flipped) selected versions, and {\em fully flipped} means every selected version is misoriented (flipped).}
\label{fig:boxoffice-cc-conditioning}
\end{figure}

Figure~\ref{fig:boxoffice-cc-conditioning} provides additional insights on the accuracy dynamics (top) for warm online methods, describing how additional versions available for the KV cache of a chunk may change the configuration of a query from aligned to nonaligned/mixed (left) and from nonaligned to mixed/aligned.

For the aligned case, when $M>1$ it suffices that a single KV cache is picked with the wrong warmup conditioning to shift a query from aligned to mixed. With 10 chunks in a query, even $M=2$ suffices to break the perfect alignment of more than $75\%$ of the queries, which explains the sudden drop in accuracy.
\looseness=-1 For the nonaligned case the dual happens: having more versions to pick from enables CacheCraft to turn some nonaligned queries to mixed ones, which partially recovers accuracy.

\section{Reuse characterization}\label{appendix:plots}
\begin{figure}[htbp]
\centering
\includegraphics[width=\linewidth]{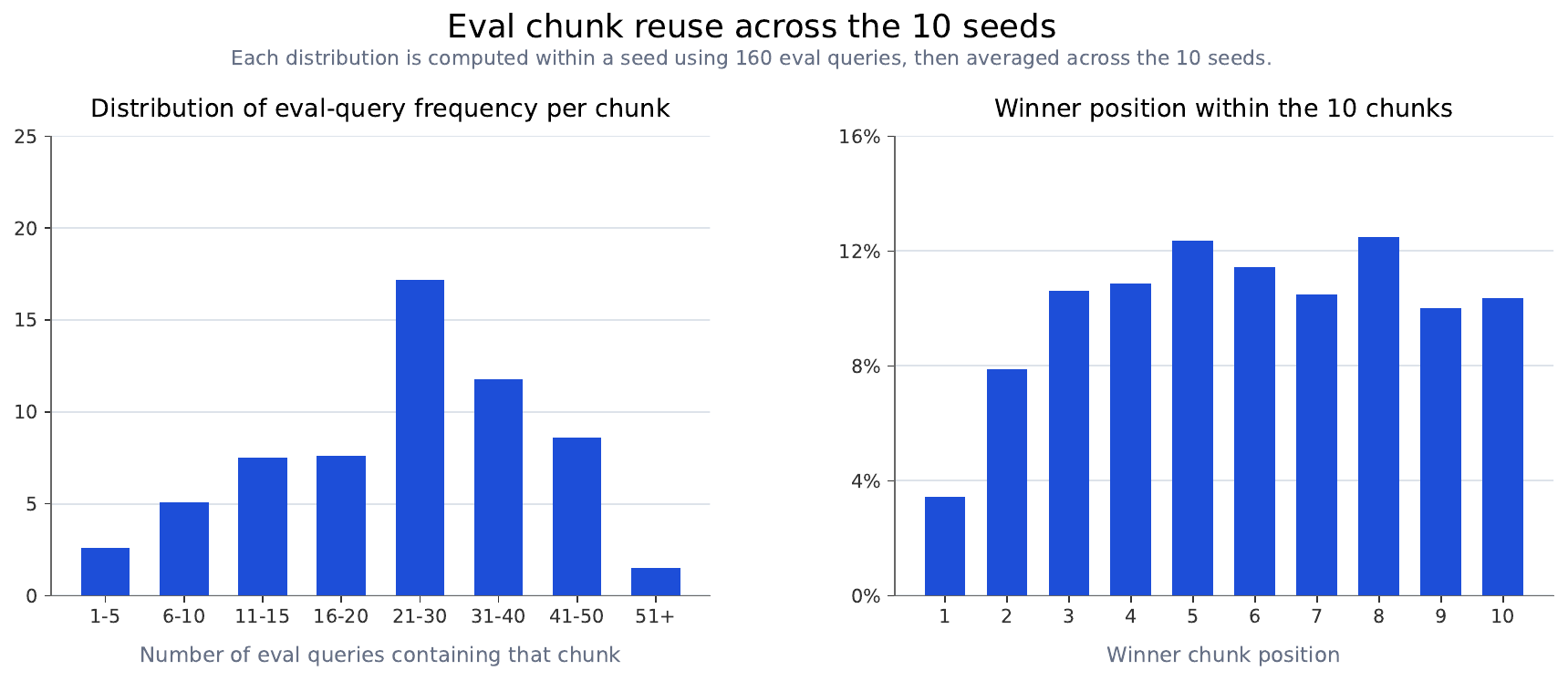}
\caption{Evaluation queries chunk reuse structure on ten \bo{} seeds. The left panel shows the distribution of eval query frequencies per unique chunk, split into winner chunks and nonwinner chunks.  The right panel shows the winner's position among the 10 candidate chunks.}
\label{fig:boxoffice-reuse-structure}
\end{figure}
Figure~\ref{fig:boxoffice-reuse-structure} provides additional information about the reuse of the evaluation chunks. For each seed, chunk frequency is computed from its 160 evaluation queries, and the plotted distributions are then averaged across seeds. 
Figure~\ref{fig:boxoffice-reuse-structure} (left)  shows for each unique chunk in the evaluation set, how many distinct evaluation queries contain it: the $x$-axis bins chunks by their evaluation-query frequency, and the $y$-axis counts how many unique chunks fall in each bin. The plot shows that there is a high degree of reuse of chunks. The plot does not reflect the warmup queries, which already guarantee, in our experiments, that each evaluation-chunk appears at least $M=4$ times.
Figure~\ref{fig:boxoffice-reuse-structure} (right) shows that winner chunks are placed at arbitrary  positions in the prompts so as to avoid positional bias.

\section{Offline warmup details}
\begin{figure}[htbp]
\centering
\includegraphics[width=\linewidth]{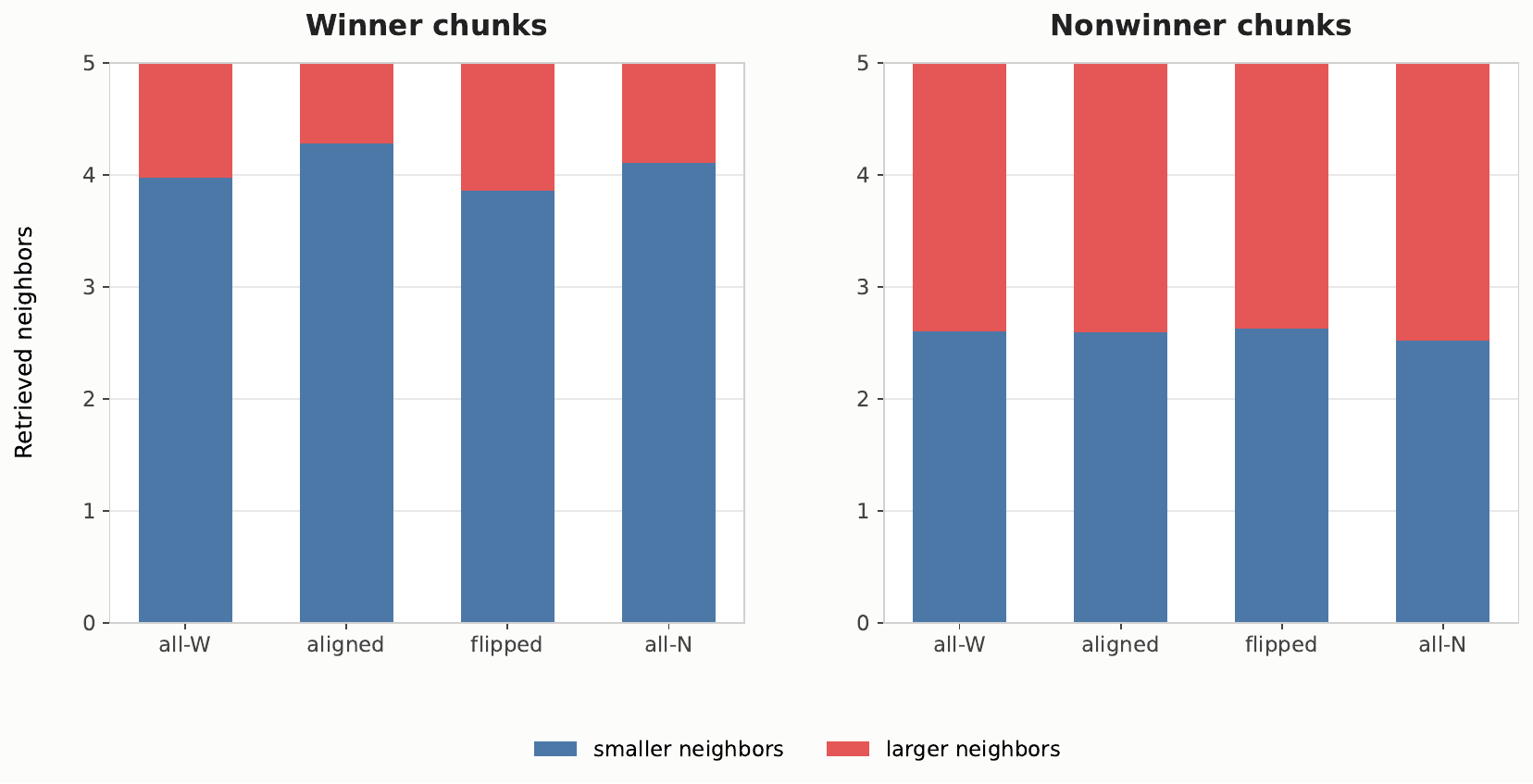}
\caption{Warm offline retrieval conditioning composition on the ten seeds \bo{} data as done in $FR$. The stacked bars show the average composition of the retrieved warm neighbors for $k{=}5$, split into smaller versus larger box-office movies relative to the target chunk. Winner chunks by construction have relatively high box office value, so they retrieve mostly smaller neighbors in every configuration. Nonwinner chunks have a more balanced neighborhood. In both cases, the KV cache encodes mixed information about the ranking of a chunk  with respect to its neighbors.}
\label{fig:boxoffice-warm-conditioning}
\end{figure}

Figure~\ref{fig:boxoffice-warm-conditioning} provides details on the conditioning of KV caches in $FR$. Both winners and nonwinners are conditioned to a mix of chunks with larger and smaller box office values, so $FR$ has no clear advantages in any query configurations.

\section{Sensitivity to the no-context threshold and size of the meaningful subset}\label{app:tau}
The NC filter of Section~\ref{sec:limitations:methodology} classifies a query as answerable without context when the model scores $F_1 > \tau$ on the question alone, with $\tau{=}0.2$ in the main text. Table~\ref{tab:app:tau} reports $\rho_{F_1}$, averaged over the nine dataset$\times$model cells, for $\tau$ between $0$ and $0.5$. $\rho$ remains well below $1$ for every method at every threshold, and grows slowly with $\tau$ because a looser threshold filters out fewer queries, so the filtered and unfiltered datasets overlap more. The conclusion of Section~\ref{sec:limitations:methodology}, that measuring $F_1$ on the unfiltered dataset inflates the reported accuracy preservation, therefore does not depend on the choice of $\tau$.

Table~\ref{tab:app:subset} reports how many of the $200$ queries of each LongBench-QA dataset survive each filter, per model. The meaningful subset is model-dependent by construction: accuracy preservation is defined relative to a specific model's full-prefill ceiling, so a query that one model answers from parametric memory may be meaningful for another. All cross-method comparisons in this paper are within-model and are therefore unaffected; to compare across models on a common set, one can take the intersection of the per-model meaningful subsets. \bo{} sidesteps the issue: its generator retains only the queries that every filter model answers with full context and none answers without it (Algorithm~1, steps 9--12), so each released dataset is a single evaluation set shared by all models, and generating a set that is meaningful for a new group of models is fully automated.

\begin{table}[t]
\centering\footnotesize\setlength{\tabcolsep}{4pt}
\caption{Sensitivity of $\rho_{F_1}$ (mean over the $9$ dataset$\times$model cells, $R{=}0.15$) to the no-context threshold $\tau$ that defines the NC filter. Left: after the NC filter; right: after NC and LI, i.e., on the meaningful subset. The row for $B_0$ alone does not depend on $\tau$: CB 0.87, CB+Q 0.85, FR 0.83, FR+Q 0.83. The paper uses $\tau{=}0.2$.}
\label{tab:app:tau}
\begin{tabular}{c cccc c cccc}
\toprule
 & \multicolumn{4}{c}{$+$NC} & & \multicolumn{4}{c}{$+$LI (meaningful subset)} \\
\cmidrule(lr){2-5}\cmidrule(lr){7-10}
$\tau$ & CB & CB+Q & FR & FR+Q & & CB & CB+Q & FR & FR+Q \\
\midrule
0.00 & 0.81 & 0.74 & 0.76 & 0.78 & & 0.80 & 0.72 & 0.73 & 0.77 \\
0.05 & 0.81 & 0.74 & 0.76 & 0.78 & & 0.80 & 0.72 & 0.73 & 0.77 \\
0.10 & 0.82 & 0.74 & 0.77 & 0.78 & & 0.80 & 0.73 & 0.73 & 0.77 \\
0.15 & 0.82 & 0.75 & 0.76 & 0.78 & & 0.81 & 0.73 & 0.74 & 0.77 \\
0.20 (paper) & 0.82 & 0.75 & 0.77 & 0.78 & & 0.81 & 0.73 & 0.74 & 0.77 \\
0.25 & 0.82 & 0.76 & 0.77 & 0.79 & & 0.81 & 0.74 & 0.75 & 0.79 \\
0.30 & 0.82 & 0.76 & 0.78 & 0.79 & & 0.81 & 0.74 & 0.75 & 0.79 \\
0.35 & 0.82 & 0.77 & 0.78 & 0.80 & & 0.82 & 0.76 & 0.76 & 0.79 \\
0.40 & 0.82 & 0.77 & 0.78 & 0.80 & & 0.81 & 0.76 & 0.76 & 0.79 \\
0.50 & 0.83 & 0.78 & 0.80 & 0.81 & & 0.82 & 0.78 & 0.78 & 0.81 \\
\bottomrule
\end{tabular}
\end{table}

\begin{table}[t]
\centering\footnotesize\setlength{\tabcolsep}{5pt}
\caption{Number of queries that survive each filter on LongBench-QA ($200$ queries per dataset). $B_0$ is independent of $\tau$; $+$NC and $+$LI are reported at the paper's $\tau{=}0.2$, and the size of the meaningful subset ($+$LI) is additionally reported for $\tau\in\{0.1,0.3,0.5\}$.}
\label{tab:app:subset}
\begin{tabular}{ll r r rr rrr}
\toprule
 & & & & \multicolumn{2}{c}{$\tau{=}0.2$} & \multicolumn{3}{c}{$+$LI at other $\tau$} \\
\cmidrule(lr){5-6}\cmidrule(lr){7-9}
dataset & model & all & $B_0$ & $+$NC & $+$LI & $0.1$ & $0.3$ & $0.5$ \\
\midrule
MuSiQue & Qwen & 200 & 79 & 47 & 47 & 44 & 52 & 69 \\
 & Llama & 200 & 79 & 57 & 57 & 52 & 60 & 70 \\
 & Mistral & 200 & 68 & 55 & 55 & 55 & 55 & 64 \\
\midrule
2WikiMQA & Qwen & 200 & 96 & 42 & 38 & 37 & 41 & 51 \\
 & Llama & 200 & 114 & 65 & 46 & 43 & 50 & 57 \\
 & Mistral & 200 & 127 & 81 & 58 & 57 & 59 & 67 \\
\midrule
HotpotQA & Qwen & 200 & 136 & 68 & 65 & 64 & 68 & 84 \\
 & Llama & 200 & 134 & 70 & 65 & 60 & 66 & 79 \\
 & Mistral & 200 & 127 & 80 & 74 & 73 & 75 & 84 \\
\bottomrule
\end{tabular}
\end{table}

\section{Larger models: Qwen3-32B}\label{app:32b}
\mypar{Accuracy inflation.} To check whether the inflation documented in Section~\ref{sec:limitations:methodology} survives at larger scale, we re-ran the evaluation of Table~\ref{tab:pitfalls} with Qwen3-32B on MuSiQue. Table~\ref{tab:app:32b-lb} reports $\rho_{F_1}$ and the $B_0$ ablation. Out of $200$ queries, $100$ are answerable with full prefill, $68$ of them are also grounded, and none is low-information, so the meaningful subset has $68$ queries. $\rho_{F_1}$ stays below $1$ for every method, at $\tau{=}0.2$ as well as at $\tau{=}0.5$, and the share of the aggregate $F_1$ that comes from baseline-zero queries ranges from $9\%$ to $24\%$. The inflation is attenuated with respect to the $8$B models (compare with the MuSiQue rows of Table~\ref{tab:pitfalls}), as expected: a larger model answers correctly from the context a larger fraction of the queries, so fewer unanswerable rows remain to inflate the metric. The inflation is thus not an artifact of small models; it persists at $4\times$ the scale.

\begin{table}[t]
\centering\footnotesize\setlength{\tabcolsep}{5pt}
\caption{Qwen3-32B on MuSiQue at $R{=}0.15$ ($200$ queries; $100$ survive $B_0$, $68$ survive $+$NC and $+$LI at $\tau{=}0.2$). Left: $\rho_{F_1}$ under the progressive filters at $\tau{=}0.2$ and at $\tau{=}0.5$. Right: $B_0$ ablation, as in Table~\ref{tab:pitfalls}.}
\label{tab:app:32b-lb}
\begin{tabular}{l ccc cc c rrr}
\toprule
 & \multicolumn{3}{c}{$\tau{=}0.2$} & \multicolumn{2}{c}{$\tau{=}0.5$} & & \multicolumn{3}{c}{$B_0$ ablation} \\
\cmidrule(lr){2-4}\cmidrule(lr){5-6}\cmidrule(lr){8-10}
method & $B_0$ & $+$NC & $+$LI & $+$NC & $+$LI & & $B_0M_0$ & $B_0M_+$ & $\%F_1$ \\
\midrule
CB   & 0.86 & 0.82 & 0.82 & 0.82 & 0.82 & & 82 & 18 & 13.8 \\
CB+Q & 0.76 & 0.72 & 0.72 & 0.71 & 0.71 & & 73 & 27 & 23.6 \\
FR   & 0.91 & 0.86 & 0.86 & 0.89 & 0.89 & & 81 & 19 & 9.3 \\
FR+Q & 0.88 & 0.81 & 0.81 & 0.85 & 0.85 & & 76 & 24 & 11.7 \\
\bottomrule
\end{tabular}
\end{table}

\mypar{Cache staleness.} We also ran the \bo{} protocol of Section~\ref{sec:boxoffice:results} with Qwen3-32B (Table~\ref{tab:app:32b-bo}). The staleness dynamic is unchanged at this scale: on \texttt{aligned}, $LM$ matches the full-prefill baseline (norm-F1 $\geq 1$), on \texttt{flipped} it scores $0.00$, and the mixed configurations sit close to \texttt{flipped}, while the cold method $CB$ is flat across configurations. Multi-version caching with $M{=}4$ lifts the three non-aligned configurations ($0.33$--$0.42$ for $CC$-$M4$ and $CC$-$M4{+}Q$, versus $0.00$--$0.11$ for $LM$) at the cost of the aligned one, mirroring Result~4 of Section~\ref{sec:boxoffice:results}; query-driven token selection helps $LM$ marginally and hurts the cold method on this model, mirroring Result~3.

\begin{table}[t]
\centering\footnotesize\setlength{\tabcolsep}{5pt}
\caption{norm-F1 of Qwen3-32B on the \bo{} argmax template at $R{=}0.15$, per warmup configuration. CB and LM are averaged over 6 generated datasets (seeds $7,11,13,17,19,23$); the remaining methods over the 3 datasets on which they were run (seeds $7,11,13$). FR is not reported because its offline pool was not built for this model. The datasets were filtered with the three $8$B models, so the Qwen3-32B full-prefill baseline is not exactly $1$ (mean $F_1$ between $0.94$ and $0.98$ per dataset) and norm-F1 can slightly exceed $1$.}
\label{tab:app:32b-bo}
\begin{tabular}{lcccccccc}
\toprule
Config & CB & CB+Q & LM & LM+Q & CC-M2 & CC-M2+Q & CC-M4 & CC-M4+Q \\
\midrule
aligned & .26 & .08 & 1.03 & \textbf{1.04} & .10 & .13 & .19 & .27 \\
all-W & .28 & .07 & .11 & .14 & .21 & .21 & .39 & \textbf{.42} \\
all-N & .35 & .05 & .08 & .09 & .16 & .19 & \textbf{.37} & .33 \\
flipped & .33 & .05 & .00 & .00 & .14 & .15 & \textbf{.41} & .34 \\
\midrule
average & .31 & .06 & .31 & .32 & .15 & .17 & .34 & \textbf{.34} \\
\bottomrule
\end{tabular}
\end{table}

\section{Additional query templates}\label{app:templates}
The main text evaluates the default \bo{} template, an argmax over a numeric field. To verify that the dynamics we report are not specific to numeric comparison, we implemented two further templates over the same fictional corpus and with the same generation pipeline (joint three-model filtering, $40$ evaluation queries per configuration, warmup balanced to $M{=}4$); the generators are released in the \texttt{templates\_ooak\_pj} folder of the code repository and the generated datasets in the \texttt{templates} folder of the dataset release~\citep{boxofficedata,boxofficerepo}.

\mypar{Genre uniqueness.} The query asks to identify the only candidate whose genre is not shared by any other candidate in the context. Answering requires equality matching across all candidates and uniqueness detection, rather than a numeric comparison. The warmup roles are defined by genre: a chunk cached after predecessors none of which shares its genre carries the signal ``I am unique among my predecessors'', and a chunk cached after at least one same-genre predecessor carries the opposite signal. With $g$ the unique-genre candidate, the four configurations are: \texttt{aligned}, $g$ was cached with no same-genre predecessor and every other candidate was cached with at least one same-genre predecessor; \texttt{flipped}, the opposite; \texttt{all-W}, every candidate was cached with no same-genre predecessor; \texttt{all-N}, every candidate was cached with at least one same-genre predecessor.

\mypar{Pivot join.} The query names a \emph{pivot} movie in the context and asks for the other candidate directed by the same director; exactly one such candidate exists. This is a strictly two-hop task: the model must locate the pivot's dossier, read its director, and then find the candidate that matches it, so the second step depends on what is extracted in the first. The golden chunks are the pivot and its match, and a chunk cached after a same-director predecessor carries the signal that a match exists: \texttt{aligned}, pivot and match were both cached with at least one same-director predecessor and every other candidate was cached with none; \texttt{flipped}, the opposite; \texttt{all-W}, every candidate was cached with at least one same-director predecessor; \texttt{all-N}, every candidate was cached with none.

\mypar{Results.}\looseness=-1 Tables~\ref{tab:app:ooak} and~\ref{tab:app:pj} report norm-F1 at $R{=}0.15$ on four generated datasets per template, for the same methods as Table~\ref{tab:bo-15}. Both templates are intrinsically harder than the argmax one, so absolute values are lower, but the dynamics of Section~\ref{sec:boxoffice:results} carry over. The staleness dynamic is reproduced: $LM$ peaks on \texttt{aligned} ($0.51$--$0.66$ on genre uniqueness, $0.24$--$0.42$ on pivot join) and collapses on \texttt{flipped} ($0.00$--$0.05$ and $0.01$--$0.06$), whereas the cold method $CB$ is flat across configurations, since its caches carry no warmup conditioning. Multi-version reuse ($CC$-$M2$, $CC$-$M4$) progressively recovers from the collapse, and query-driven token selection ($+Q$) improves every warm method on every model on average. Finally, as in Result~5 of Section~\ref{sec:boxoffice:results}, the best method differs across models.

\begin{table}[t]
\centering\scriptsize\renewcommand{\arraystretch}{0.95}\setlength{\tabcolsep}{1pt}
\caption{norm-F1 on the \bo{} \emph{genre-uniqueness} template at $R{=}0.15$, mean over 4 generated datasets (seeds $7,11,13,17$; $40$ evaluation queries per configuration per seed). Within each (model, configuration), the winning method is in bold.}
\label{tab:app:ooak}
\begin{tabular}{l|*{10}{c} @{\hspace{10pt}} *{10}{c} @{\hspace{10pt}} *{10}{c}}
\toprule
 & \multicolumn{10}{c}{\textbf{Llama}} & \multicolumn{10}{c}{\textbf{Qwen}} & \multicolumn{10}{c}{\textbf{Mistral}} \\
\cmidrule(lr){2-11} \cmidrule(lr){12-21} \cmidrule(lr){22-31}
Config & \rotatebox{90}{\,CB} & \rotatebox{90}{\,CB+Q} & \rotatebox{90}{\,FR} & \rotatebox{90}{\,FR+Q} & \rotatebox{90}{\,LM} & \rotatebox{90}{\,LM+Q} & \rotatebox{90}{\,CC-M2} & \rotatebox{90}{\,CC-M2+Q} & \rotatebox{90}{\,CC-M4} & \rotatebox{90}{\,CC-M4+Q} & \rotatebox{90}{\,CB} & \rotatebox{90}{\,CB+Q} & \rotatebox{90}{\,FR} & \rotatebox{90}{\,FR+Q} & \rotatebox{90}{\,LM} & \rotatebox{90}{\,LM+Q} & \rotatebox{90}{\,CC-M2} & \rotatebox{90}{\,CC-M2+Q} & \rotatebox{90}{\,CC-M4} & \rotatebox{90}{\,CC-M4+Q} & \rotatebox{90}{\,CB} & \rotatebox{90}{\,CB+Q} & \rotatebox{90}{\,FR} & \rotatebox{90}{\,FR+Q} & \rotatebox{90}{\,LM} & \rotatebox{90}{\,LM+Q} & \rotatebox{90}{\,CC-M2} & \rotatebox{90}{\,CC-M2+Q} & \rotatebox{90}{\,CC-M4} & \rotatebox{90}{\,CC-M4+Q} \\
\midrule
aligned & .69 & .40 & .08 & .31 & .51 & .68 & .35 & .65 & .33 & \textbf{.72} & .27 & .09 & .11 & .34 & .56 & \textbf{.65} & .21 & .36 & .30 & .35 & .23 & .07 & .10 & .14 & .66 & \textbf{.72} & .29 & .42 & .35 & .43 \\
all-W & \textbf{.72} & .49 & .10 & .31 & .15 & .60 & .19 & .65 & .31 & .61 & .25 & .12 & .15 & \textbf{.37} & .11 & .15 & .18 & .29 & .32 & .28 & .26 & .09 & .14 & .19 & .20 & .31 & .22 & .28 & .36 & \textbf{.42} \\
all-N & .68 & .38 & .08 & .23 & .13 & .66 & .28 & .66 & .35 & \textbf{.70} & .34 & .17 & .12 & \textbf{.42} & .11 & .39 & .26 & .34 & .31 & .39 & .24 & .08 & .12 & .18 & .18 & .27 & .26 & .36 & .40 & \textbf{.45} \\
flipped & .64 & .44 & .07 & .24 & .05 & .52 & .19 & \textbf{.65} & .30 & .59 & .25 & .16 & .13 & .30 & .00 & .09 & .18 & .31 & \textbf{.32} & \textbf{.32} & .26 & .08 & .09 & .17 & .04 & .08 & .26 & .33 & .36 & \textbf{.37} \\
\midrule
average & \textbf{.68} & .43 & .08 & .27 & .21 & .61 & .25 & .65 & .33 & .66 & .28 & .13 & .13 & \textbf{.36} & .20 & .32 & .21 & .33 & .31 & .33 & .24 & .08 & .11 & .17 & .27 & .35 & .26 & .35 & .37 & \textbf{.42} \\
\bottomrule
\end{tabular}
\end{table}

\begin{table}[t]
\centering\scriptsize\renewcommand{\arraystretch}{0.95}\setlength{\tabcolsep}{1pt}
\caption{norm-F1 on the \bo{} \emph{pivot-join} template at $R{=}0.15$, mean over 4 generated datasets (seeds $7,11,13,17$; $40$ evaluation queries per configuration per seed). Within each (model, configuration), the winning method is in bold.}
\label{tab:app:pj}
\begin{tabular}{l|*{10}{c} @{\hspace{10pt}} *{10}{c} @{\hspace{10pt}} *{10}{c}}
\toprule
 & \multicolumn{10}{c}{\textbf{Llama}} & \multicolumn{10}{c}{\textbf{Qwen}} & \multicolumn{10}{c}{\textbf{Mistral}} \\
\cmidrule(lr){2-11} \cmidrule(lr){12-21} \cmidrule(lr){22-31}
Config & \rotatebox{90}{\,CB} & \rotatebox{90}{\,CB+Q} & \rotatebox{90}{\,FR} & \rotatebox{90}{\,FR+Q} & \rotatebox{90}{\,LM} & \rotatebox{90}{\,LM+Q} & \rotatebox{90}{\,CC-M2} & \rotatebox{90}{\,CC-M2+Q} & \rotatebox{90}{\,CC-M4} & \rotatebox{90}{\,CC-M4+Q} & \rotatebox{90}{\,CB} & \rotatebox{90}{\,CB+Q} & \rotatebox{90}{\,FR} & \rotatebox{90}{\,FR+Q} & \rotatebox{90}{\,LM} & \rotatebox{90}{\,LM+Q} & \rotatebox{90}{\,CC-M2} & \rotatebox{90}{\,CC-M2+Q} & \rotatebox{90}{\,CC-M4} & \rotatebox{90}{\,CC-M4+Q} & \rotatebox{90}{\,CB} & \rotatebox{90}{\,CB+Q} & \rotatebox{90}{\,FR} & \rotatebox{90}{\,FR+Q} & \rotatebox{90}{\,LM} & \rotatebox{90}{\,LM+Q} & \rotatebox{90}{\,CC-M2} & \rotatebox{90}{\,CC-M2+Q} & \rotatebox{90}{\,CC-M4} & \rotatebox{90}{\,CC-M4+Q} \\
\midrule
aligned & \textbf{.62} & .59 & .09 & .21 & .24 & .52 & .15 & .40 & .23 & .46 & .50 & .29 & .14 & .29 & .42 & \textbf{.56} & .27 & .47 & .29 & .46 & .39 & .08 & .19 & .25 & .28 & \textbf{.56} & .24 & \textbf{.56} & .23 & .49 \\
all-W & \textbf{.60} & .56 & .12 & .20 & .15 & .39 & .18 & .45 & .23 & .53 & \textbf{.52} & .26 & .15 & .23 & .14 & .21 & .28 & .49 & .31 & .43 & .42 & .08 & .14 & .26 & .09 & .31 & .31 & .54 & .36 & \textbf{.55} \\
all-N & .67 & \textbf{.68} & .14 & .23 & .10 & .38 & .14 & .49 & .14 & .51 & \textbf{.48} & .25 & .06 & .27 & .10 & .44 & .22 & .42 & .29 & .44 & .43 & .08 & .16 & .31 & .13 & .46 & .22 & \textbf{.48} & .35 & \textbf{.48} \\
flipped & \textbf{.63} & .59 & .22 & .23 & .06 & .34 & .19 & .52 & .21 & .54 & .47 & .31 & .11 & .21 & .01 & .08 & .13 & .32 & .22 & \textbf{.50} & .46 & .08 & .16 & .31 & .03 & .32 & .19 & .52 & .39 & \textbf{.56} \\
\midrule
average & \textbf{.63} & .61 & .14 & .22 & .14 & .40 & .17 & .46 & .21 & .51 & \textbf{.49} & .28 & .11 & .25 & .17 & .32 & .22 & .43 & .28 & .46 & .42 & .08 & .17 & .28 & .13 & .41 & .24 & \textbf{.52} & .33 & \textbf{.52} \\
\bottomrule
\end{tabular}
\end{table}

\section{Evidence from production RAG workloads}\label{app:traces}
\bo{} generates, by design, workloads in which the same chunk recurs across queries while its context differs at every recurrence. We checked whether this regime is representative of deployed RAG systems using the industry-oriented data we are aware of: RAGPulse~\citep{ragpulse}, a production trace of a deployed RAG service; WixQA~\citep{wixqa}, real enterprise support queries over a shared knowledge base; HERB~\citep{herb}, an industry-authored enterprise-workflow benchmark; and the production-workload analysis published with Cache-Craft~\citep{cachecraft}, whose underlying data is not released.

All four indicate that chunk reuse is pervasive. In RAGPulse, $97\%$ of the retrieved-passage slots land on passages used by at least two requests; in WixQA, popular articles serve up to $14$ distinct user queries; HERB exhibits pervasive cross-question reuse of golden evidence chunks; Cache-Craft reports that $75\%$ of the chunks retrieved per query are reprocessed ones and that the top $5\%$ of the chunks serve $60\%$ of the requests. On the three multi-chunk resources (RAGPulse, about $6$ retrieved passages per request; HERB, about $9$ evidence chunks per question; Cache-Craft, $5$--$15$ chunks per query), the reuse is not clustered: the same chunk keeps returning, but its context differs at every recurrence. In RAGPulse, requests that share a chunk have a median Jaccard similarity of $0.09$ between their chunk sets; since RAGPulse releases only chunk hashes, the offline-similarity assumption of Section~\ref{sec:limitations:offline} cannot be tested on it. In HERB, where the text is available, the Jaccard similarity between questions that share a chunk is likewise about $0.1$, and we tested the offline-similarity assumption directly: after embedding all $8{,}348$ chunks, a chunk's actually co-used partner is among its top-$10$ offline nearest neighbors in only $7.9\%$ of the $54{,}974$ co-used pairs. Cache-Craft reports the same from the systems side: chunks recur constantly but identical combinations rarely do (the chunks of a query typically come from three or more past requests), and its production deployment stores up to $11$ differently-contextualized cache variants of a single chunk. WixQA is excluded from this second analysis by construction: its answers draw on $1.15$--$1.29$ articles on average, a near-single-chunk regime in which cross-chunk composition barely arises.

Production systems thus handle chunks that are highly prone to reuse and that are combined at query time in ways that are hard to predict offline. This is exactly the regime that \bo{} generates by design: on the full \bo{} workload (about $250$ warmup and $160$ evaluation queries per seed), $100\%$ of chunk slots land on chunks used by at least two queries, and queries that share a chunk have a median Jaccard similarity of $0.11$ in every seed, close to RAGPulse's $97\%$ and $0.09$. The number of cached versions $M$, the mix of warmup configurations, the candidate-set size, and the recomputation budget $R$ are settable to match such measured statistics.

\end{document}